\documentclass[sigconf,authorversion]{acmart}

\usepackage{booktabs} % For formal tables
\usepackage[utf8]{inputenc} % allow utf-8 input
\usepackage[T1]{fontenc}    % use 8-bit T1 fonts
\usepackage{url}            % simple URL typesetting
\usepackage{booktabs}       % professional-quality tables
\usepackage{amsfonts}       % blackboard math symbols
\usepackage{nicefrac}       % compact symbols for 1/2, etc.
\usepackage{microtype}      % microtypography
\usepackage{amsmath}
\usepackage{graphicx}
\usepackage{subcaption}
\usepackage{bm}
\usepackage[title]{appendix}
\usepackage{verbatim}

\usepackage{EvolvingAIcommenting}
\newif\ifcomments

\commentstrue

\ifcomments
\newcommand{\comments}[1]{#1}
\else
\newcommand{\comments}[1]{}
\fi
\setcopyright{rightsretained}
\copyrightyear{2018} 
\acmYear{2018} 
\acmConference[GECCO '18]{Genetic and Evolutionary Computation Conference}{July 15--19, 2018}{Kyoto, Japan}
\acmPrice{15.00}
\acmDOI{10.1145/3205455.3205473}
\acmISBN{978-1-4503-5618-3/18/07}

\begin{document}
\title{Safe Mutations for Deep and Recurrent Neural Networks through Output Gradients}
\author{Joel Lehman, Jay Chen, Jeff Clune, and Kenneth O. Stanley}
\affiliation{%
  \institution{Uber AI Labs, San Francisco, CA}
  %\streetaddress{1455 Market Street}
  %\city{San Francisco} 
  %\state{CA} 
  %\postcode{}
}
\email{ {joel.lehman,jayc,jeffclune,kstanley}@uber.com}
\begin{comment}
\author{Joel Lehman}
\affiliation{%
  \institution{Uber AI Labs}
  %\streetaddress{1455 Market Street}
  \city{San Francisco} 
  \state{CA} 
  %\postcode{}
}
\email{joel.lehman@uber.com}

\author{Jay Chen}
\affiliation{%
  \institution{Uber AI Labs}
    \city{San Francisco} 
      \state{CA} 
}
\email{jayc@uber.com}

\author{Jeff Clune}
\affiliation{%
  \institution{Uber AI Labs}
    \city{San Francisco} 
      \state{CA} 
}
\email{jeffclune@uber.com}

\author{Kenneth O. Stanley}
\affiliation{%
  \institution{Uber AI Labs}
      \city{San Francisco} 
      \state{CA} 
}
\email{kstanley@uber.com}
\end{comment}


%KENTODO: note I slightly altered the abstract so we should update it on the GECCO submission site
\begin{abstract}
While \emph{neuroevolution} (evolving neural networks) has been successful across a variety of domains from reinforcement learning, to artificial life, to evolutionary robotics, it is rarely applied to large, deep neural networks. A central reason is that while random mutation generally works in low dimensions, a random perturbation of thousands or millions of weights will likely break existing functionality. This paper proposes a solution: a family of \emph{safe mutation} (SM) operators that facilitate exploration without dramatically altering network behavior or requiring additional interaction with the environment.  The most effective SM variant scales the degree of mutation of each individual weight according to the sensitivity of the network's outputs to that weight, which requires computing the gradient of outputs with respect to the weights (instead of the gradient of error, as in conventional deep learning).  This \emph{safe mutation through gradients} (SM-G) operator dramatically increases the ability of a simple genetic algorithm-based neuroevolution method to find solutions in high-dimensional domains that require deep and/or recurrent neural networks, including domains that require processing raw pixels.  By improving our ability to evolve deep neural networks, this new safer approach to mutation expands the scope of domains amenable to neuroevolution. 
\end{abstract}

\begin{CCSXML}
<ccs2012>
<concept>
<concept_id>10010147.10010257.10010293.10011809.10011812</concept_id>
<concept_desc>Computing methodologies~Genetic algorithms</concept_desc>
<concept_significance>500</concept_significance>
</concept>
<concept>
<concept>
<concept_id>10010147.10010257.10010293.10010294</concept_id>
<concept_desc>Computing methodologies~Neural networks</concept_desc>
<concept_significance>300</concept_significance>
</concept>
<concept>
<concept_id>10010147.10010257.10010293.10011809.10011810</concept_id>
<concept_desc>Computing methodologies~Artificial life</concept_desc>
<concept_significance>300</concept_significance>
</concept>
<concept>
<concept_id>10010147.10010257.10010293.10011809.10011814</concept_id>
<concept_desc>Computing methodologies~Evolutionary robotics</concept_desc>
<concept_significance>300</concept_significance>
</concept>
</ccs2012>
\end{CCSXML}

\vspace{-0.05in}
\ccsdesc[500]{Computing methodologies~Genetic algorithms}
\ccsdesc[300]{Computing methodologies~Neural networks}
\ccsdesc[300]{Computing methodologies~Artificial life}
\ccsdesc[300]{Computing methodologies~Evolutionary robotics}

\vspace{-0.05in}
\keywords{Neuroevolution, mutation, deep learning, recurrent networks}

\maketitle

%TODO:
%Include source code in supplemental material
%--Remove youtube playlist for review...

\vspace{-0.05in}
\section{Introduction}

Neuroevolution (NE; \citep{yao:ieee99}) combines evolutionary algorithms (EAs) and artificial neural networks (NNs).  It has long been popular when gradient information with respect to the task is unavailable or difficult to obtain, e.g.\ in artificial life \citep{langton:artificial} and evolutionary robotics \citep{nolfi:book00}. Interestingly for NE, recent advances in deep learning have demonstrated the value of large NNs \citep{goodfellow:deep}, which could seemingly open up new horizons for NE if only it could scale to effectively evolve NNs with millions of parameters (although recent work suggests this may sometimes already be possible \citep{es,such:arxiv17}). However, NE historically has operated on much smaller networks, most often on the order of tens or hundreds of parameters. %due to at least the perception of the so-called curse of dimensionality, as well as more limited computational resources than today. 

The challenge is that large NNs induce a tension between how fast evolution can theoretically proceed and the degree of weight perturbations applied as a mutation operator. If few weights are perturbed in each generation, it would take many generations for all weights to be tuned; but if many weights are perturbed, changes to the NN's functionality may be too drastic for search to proceed systematically. 
Indeed, such concerns inspired research into indirect encodings \citep{stanley:taxonomy}, wherein a compact genotype is expanded at evaluation-time into a larger NN phenotype. 
%EMAILED - KENTODO: I don't agree with the sentence below about the problem with indirect encoding and in my view it perpetuates a myth.  What about this alternative: "While such indirect encodings have enable evolving large NNs \citep{stanley:alife09,van:wavelet}, even the indirect encoding itself can become high-dimensional, and we would ideally like to be able to efficiently evolve genomes of any dimensionality regardless of the encoding."  We can discuss it if you want, but if you're okay with it then let's switch it in.
While such indirect encodings have enabled evolving large NNs \citep{stanley:alife09,van:wavelet}, even the indirect encoding itself can become high-dimensional and thereby complicate evolution.
%TODO(KEN): Any non-hyperneat large NE indirect encoding examples? [joelnote: added a schmidhuber wavelet encoding that evolves nets with 800k weights]
A further challenge is that in deep and recurrent NNs, there may be drastic differences in the \emph{sensitivity} of parameters; for example, perturbing a weight in the layer immediately following the inputs has rippling consequences as activation propagates through subsequent layers, unlike perturbation in layers nearer to the output. In other words, simple mutation schemes are unlikely to tune individual parameters according to their sensitivity. 

To address these challenges with mutating large and deep NNs, this paper pursues the largely unexplored approach of considering perturbation in the space of an NN's \emph{outputs} rather than only in the space of its \emph{parameters}. By considering the NN's structure and the context of past inputs and outputs, it becomes possible to deliberately construct perturbations that avoid wrecking functionality while still encouraging exploring new behaviors. The aim is to ensure that an offspring's NN response will not diverge too drastically from the response of its parent. For example, when the NN is differentiable (which does not require that the \emph{task} or \emph{reward} is differentiable), gradient information can estimate how sensitive the NN's output is to perturbations of \emph{individual} parameters. The opportunity to exploit gradients in \emph{mutation} instead of error or reward, as in stochastic gradient descent (SGD), is an intriguing hint that the line between deep learning and neuroevolution is more blurry than may previously have been appreciated.

This insight leads to two approaches to generating safer NN mutations. One is called \emph{safe mutation through rescaling} (SM-R): At the expense of several NN forward passes, a line search can rescale the magnitude of a raw weight perturbation until it is deemed safe, which does not require the NN to be differentiable. The second is called \emph{safe mutation through gradients} (SM-G): When the NN is differentiable, the sensitivity of the NN to relevant input patterns can be calculated (at the expense of a backward pass). Importantly, the assumption underlying these approaches is that \emph{domain} evaluation (i.e.\ rollouts) is expensive relative to \emph{NN} evaluation (i.e.\ forward or backward NN propagation). Interestingly, both approaches relate to effective mechanisms from deep learning, e.g.\ adaptive-learning rate methods \citep{kingma:adam} or trust regions \citep{schulman:trpo}, although here there is distinct motivation and setting (i.e.\ SM-R and SM-G generate pure variation independent of reward, unlike such deep learning methods).

Both approaches are tested in this paper against a suite of supervised and reinforcement learning (RL) tasks, using large, deep, and recurrent NNs. Note that the EA in these experiments is a simple steady-state algorithm on high-dimensional vectors, so there is no additional sophisticated machinery at work.  
%(For example, algorithms like the ES of \citet{es} may induce their own implicit pressure towards regions of the search space where mutations are safer \citep{lehman:arxiv17fd}, but the setup in this paper receives no such additional benefits.)
The tasks build in complexity, ultimately showing their promise for scaling NE to challenging and high-dimensional domains, such as learning to play video games from raw pixels. The results show that in some domains where it would otherwise be  be impossible to effectively evolve networks with hundreds of thousands of connections or over a dozen layers (at least with a traditional EA), SM-G 
entirely changes what is possible and with little computational cost.
Furthermore, the value of gradient information is confirmed by an evident advantage of SM-G over SM-R when gradient information is available, especially in evolving large recurrent networks. The conclusion is that SM-G may help to unlock the power of large-scale NN models for NE and EAs in general.

\vspace{-0.1in}
\section{Background}

%* Background
%   * GPO paper (https://openreview.net/forum?id=ByOnmlWC-&noteId=ByOnmlWC-)
%      * Unlike here, they require more rollouts.  Still clearly related to SMOG.
%   * TRPO / Adam for indirect motivation
%   * CMA-ES / EDA / Self-adaptation (other approaches for more-informed offspring generation)
%   * Indirect encoding (another way to scale NE)

Next, sensitivity-aware deep learning, deep NE, and sensitivity-aware mutations for evolutionary computation (EC) are reviewed.

\vspace{-.05in}
\subsection{Sensitivity-aware Deep Learning}

%TRPO / Adam 
%Differentiability wrt to error (but also wrt to output)

There is awareness in deep learning that parameter sensitivity is important. For example, adaptive learning-rate optimizers like ADAM \citep{kingma:adam} in effect take smaller steps when the error gradient with respect to a particular parameter is large (i.e.\ it is a sensitive parameter), and larger steps when the error gradient is small (i.e.\ the parameter is relatively insensitive). The SM-G method proposed here can be seen as having motivation similar to ADAM, but with respect to generating \emph{variation} in an NN's outputs instead of directly \emph{reducing} error. Trust regions in policy gradient methods \citep{schulman:trpo} also bear similarity to SM-R and SM-G, in that they attempt to maximize improvement to an NN while limiting the degree of functional change to its behavior. A key difference is that SM-R and SM-G are uninformed by performance, and adapt candidate perturbations solely to attain a desired amount of NN output change. Additionally, other families of deep learning enhancements can be seen as attempting to reduce or normalize sensitivity. For example, long short-term memory (LSTM; \citep{hochreiter:lstm}) units are designed to avoid some of the instability of vanilla recurrent neural networks (RNNs). SM-R and SM-G can be seen as complementary to such architectural changes, which may also enable more consistent mutation.

\vspace{-0.05in}
\subsection{Deep Neuroevolution}

Recently there has been increased interest in NE from deep learning researchers evolving the architecture of deep NNs \citep{miikkulainen:arxiv17,fernando:pathnet}, which otherwise requires domain knowledge to engineer. This setting is a natural intersection between EC and deep learning; evolution discovers the structure of a deep network (for which gradient information is unavailable), while deep learning tunes its parameters through SGD. The aim in this paper is instead to exploit gradients to inform variation \emph{within} NE, meaning that this technique can be applied to problems difficult to formulate for deep learning (e.g.\ open-ended evolution or divergent search), or can improve upon where NE is already competitive with deep learning (e.g.\ the Deep GA of \citet{such:arxiv17}).

Also recently, \citet{es} demonstrated that with access to large-scale computation a form of evolution strategy (ES) scales surprisingly well to evolve deep NNs, although it remains unclear how to generalize such an approach to EAs as a whole, or when subject to a computational budget. Interestingly, this form of ES implicitly results in a limited form of safe mutation itself \citep{lehman:arxiv17fd}. The approach here aims to be less computationally expensive, and to generalize across EAs; results such as \citet{koutnik:dnn} and \citet{such:arxiv17} demonstrate the promise of such general deep NE, which safe mutation could potentially further catalyze. Some previous work in indirect encoding of NNs also are forms of deep NE \citep{stanley:alife09,van:wavelet}. However, indirect encoding offers its own challenges and thus it is useful to have direct encoding approaches to deep NE, such as SM-R and SM-G. Additionally, indirect encodings may equally benefit from the methods proposed here. 

\vspace{-0.13in}
\subsection{Informed Mutation in EC}

Because mutation is a critical EC operator, many approaches target it for improvement. For example, estimation of distribution algorithms (EDAs; \citep{pelikan:eda}) iteratively build and exploit probabilistic models of how promising solutions are distributed; such models can potentially capture the sensitivity of parameters when generating new individuals, although building such models is often expensive in practice. A related approach called natural evolution strategies (NES; \citep{wierstra:nes}) directly optimizes a distribution of solutions. This distributional optimization may indirectly encourage safer mutations by guiding the search to robust areas \citep{lehman:arxiv17fd} or by adaptively adjusting the variance of the distribution on a per-parameter basis through domain feedback. Related to NES and EDAs, CMA-ES learns to model pair-wise dependencies among parameters, aimed at generating productive offspring \citep{hansen:cmaes}. In contrast, the approaches proposed in this paper do not assume a formal distributional approach and attempt to measure sensitivity without interactions with the domain, allowing the paradigm to generalize to all EAs. Genetic policy optimization \citep{peng:gpo} is a recent approach that hybridizes EAs and policy gradients; in effect it applies policy gradients as a reward-optimizing variation operator for a specialized small-population EA, using additional domain rollouts and mechanisms such as imitation learning and stochastic-output NNs. (In contrast, the safe mutations in this paper require no additional domain evaluations.) %DONE -- KENTODO: Should we add "...require no additional rollouts in the domain."?  That is, "in the domain" might help remove ambiguity about whether "rollout" might just mean running the NN on its own on historical data outside the domain.

Interestingly, these mechanisms all attempt to learn from reward, which limits applying them to less-conventional EAs. For example, in EAs focused on creative divergence \citep{lehman:ecj11,mouret:arxiv15}, a population may span many non-stationary modes of high reward, thereby increasing the challenge for approaches that model and track reward distributions. Similarly, within artificial life or open-ended evolution, the concept of an overarching reward function may not even be meaningful. Other EC research explores how \emph{selection pressure} can drive search towards robust or evolvable areas of the search space \citep{lehman:plos13,wilke:nature01b} or how evolution can produce healthier variation when mutation operators can themselves vary \citep{meyer:selfadapt}. Such methods may naturally complement SM-G and SM-R, e.g.\ self-adaptation could adapt the intensity of SM's informed mutations.%EMAIL KENTODO: You say "they focus on selection," but does the meyer:selfadapt citation mentioned right before this last sentence really focus on selection?

\vspace{-0.in}
\section{Approach}

%* Approach
%   * Limited output change  (previously called TRPO)
%   * Sensitivity-aware mutation (previously called ADAM)
%   * KEN: So LOC and SAM?  What happens to SMOG?
%   * TODO: Mathematical basis for sensitivity formula

The general approach for safe mutations in this paper is to choose weight perturbations in an informed way such that they produce 
limited change in the NN's response to representative input signals. The idea is to exploit sources of information that while generally are freely available, are 
often ignored and discarded. In particular, an archive of representative \emph{experiences} and corresponding NN \emph{responses} can be gathered 
during an individual's evaluation, which can serve to ground how dramatically a weight perturbation will changes the NN's responses, and thereby inform how
its offspring are generated. Secondly, when available, knowledge about the NN structure can also be leveraged to estimate the local effect of weight perturbations on an NN's outputs (as explained later).

%notation standards: bold italics indicates vector
%				     bold non-italic indicates random vector
%					 bold caps indicates matrix

To frame the problem, consider a parent individual with parameters $\bm{w}$ and a potential parameter perturbation vector $\boldsymbol{\delta}$. While in general $\boldsymbol{\delta}$ can be sampled in many ways, here the assumption is that $\boldsymbol{\delta}$ is drawn from an isotropic normal distribution (i.e.\ the standard deviation is the same for each parameter of the NN), as in \citet{es}. When the parent is evaluated, assume that a matrix $\bm{X}_{ij}$ is recorded, consisting of the parent's $i$th sampled input experience (out of $I$ total sampled experiences) of its $j$th input neuron, along with the NN's corresponding output response, $\bm{Y}_{ik} = \textrm{NN}(\bm{X}_{i} ; \bm{w})_{k}$, to each experience (where $k$ indexes the NN's outputs). Note that $\textrm{NN}(\bm{x};\bm{w})$ represents the result of forward-propagating the input vector $\bm{x}$ through an NN with weights $\bm{w}$.

It is then possible to express how much an NN's response \emph{diverges} as a result of perturbation $\boldsymbol{\delta}$:

\vspace{-0.15in}
\begin{equation} \label{eq:divergence}
	\hspace{-0.02in}\textrm{Divergence}(\boldsymbol{\delta};\bm{w}) = \frac{\sum_{i} \sum_{k} (\textrm{NN}(\bm{X}_i;\bm{w})_k - \textrm{NN}(\bm{X}_i{;}\bm{w}+\boldsymbol{\delta})_k)^2}{I}.
\end{equation}
\vspace{-0.1in}

With this formalism, one can specify the desired amount of divergence (i.e.\ a mutation that induces some limited change), and then search for or calculate a perturbation $\boldsymbol{\delta}$ that satisfies the requirement. There are different ways to approach safe mutation from this perspective. In particular, the next section explores a simple mechanism to rescale the magnitude of a weight perturbation to encourage safe mutations when the NN is not differentiable. The following section then introduces more flexible perturbation-adjustment methods that exploit gradients through the model. Note that section 1 of the supplemental material describes experiments with a simple poorly-conditioned model that ground intuitions about specific properties of the different SM methods; these experiments are referenced below when relevant and are optional for understanding the paper as a whole. 

\vspace{-0.1in}
\subsection{Safe Mutation through Rescaling}

%One reasonable assumption is that the divergence measured by equation \ref{eq:divergence} will generally monotonically increase with the magnitude of the vector $\Delta$. This relationship will not always hold, although it does of course converge to $0$ as the magnitude of $\Delta$ approaches zero. 

%KENTODOFINAL: romanize multi-letter subscripts?
One approach to satisfying a specific level of divergence in equation \ref{eq:divergence} is called \emph{safe mutation through rescaling} (SM-R). The idea is to decompose $\boldsymbol{\delta}$ into a \emph{direction} vector in parameter space and a \emph{magnitude} scalar that specifies how far along the direction to perturb: $\boldsymbol{\delta} = \delta_{\textrm{magnitude}}\boldsymbol{\delta}_{\textrm{direction}}$. First, the direction is chosen randomly. Then, the scalar $\delta_{\textrm{magnitude}}$ is optimized with a simple line search to target a specific amount of divergence, which becomes a search hyperparameter replacing the traditional mutation rate. 

Importantly, because the parent's experiences have been recorded, this rescaling approach does not require additional domain evaluations, although it does require further NN forward passes (i.e.\ one for each iteration of the line search, assuming that the sample of experiences is small enough to fit in a single mini-batch). While this approach can achieve variation that is safe by some definition, the effects of mutation may be dominated by a few highly-sensitive parameters (see the Easy and Medium tasks in supplemental material section 1 for a toy example); in other words, this method can rescale the perturbation as a whole, but it cannot granularly rescale each dimension of a perturbation to ensure it has equal opportunity to be explored. The next section describes how gradient information can be exploited to adjust not only the magnitude, but also to reshape the direction of perturbation as well. 

\vspace{-0.1in}
\subsection{Safe Mutation through Gradients}

A more flexible way to generate safe variation is called \emph{safe mutation through gradients} (SM-G). The idea is that if the NN targeted by SM is \emph{differentiable}, then gradient information can \emph{approximate} how an NN's outputs vary as its weights are changed. In particular, the output $\bm{Y}_{ik}$ of the NN can be modeled as a function of weight perturbation $\boldsymbol{\delta}$ through the following first-order Taylor expansion:

%\begin{equation} \label{eq:taylor}
%\bm{Y}_{ij}(\bm{X_i},\boldsymbol{\delta};\bm{w}) = NN(\bm{X}_i;\bm{w})_j + \boldsymbol{\delta} \nabla_{\bm{w}}_j NN(\bm{X}{_i};\bm{w})_j + %\boldsymbol{\delta}^T H_{\bm{w}}(NN(\boldsymbol{X}_i;\boldsymbol{w}))\boldsymbol{\delta}
%\end{equation}

\vspace{-0.1in}
\begin{equation*} \label{eq:taylor}
\bm{Y}_{ik}(\bm{X_i},\boldsymbol{\delta};\bm{w}) = NN(\bm{X}_i;\bm{w})_k + \boldsymbol{\delta} \nabla_{\bm{w}} NN(\bm{X}{_i};\bm{w})_k
\end{equation*}

This approximation illustrates that the magnitude of each output's gradient with respect to any weight serves as a local estimate of the sensitivity of that output to that weight: It represents the \emph{slope} of the NN's response from perturbations of that weight. By summing such per-output weight sensitivities over outputs, the result is the overall sensitivity of a particular weight. More formally, sensitivity vector $\bm{s}$ containing the sensitivities of all weights in a NN can be calculated as:

$$\bm{s} = \sqrt[]{\sum_k \bigg(\frac{\sum_i \mathrm{abs}(\nabla_{\bm{w}} NN(\bm{X}_i)_k)}{I}\bigg)^2}.$$ 
%$$\bm{s} = \sqrt[]{\sum_k (\sum_i \mathrm{abs}(\nabla_{\bm{w}} NN(\bm{X}_i)_k))^2}.$$ 

One simple approach to adjust a perturbation on a per-parameter basis is thus to normalize a perturbation by this sensitivity: %In particular, a mini-batch of the parent's experiences $\bm{X}$ is passed through the NN, and sensitivity $S$ of weights is calculated as: 
$$\boldsymbol{\delta}_{\textrm{adjusted}} = \frac{\boldsymbol{\delta}}{\bm{s}}.$$ Note that calculating $\boldsymbol{s}$ in practice requires taking the 
average absolute value of the gradient over a batch of data (the absolute 
%KENNOTE: Maybe remove the "note that" at the start of the parenthetical - the main sentence already starts with "Note that" so it seems redundant.
value reflects that we care about the magnitude of the slope and not its sign); unfortunately this cannot be efficiently calculated within popular tensor-based machine learning platforms (e.g.\ TensorFlow or PyTorch), which are optimized to compute gradients of an aggregate scalar (e.g.\ average loss over many examples) and not aggregations over functions of gradients (e.g.\ summing the absolute value of per-example gradients). To compute this \emph{absolute gradient} variant of SM-G (\emph{SM-G-ABS}) requires a forward and backward pass for \emph{each} of the parent's experiences, which is expensive. A less-precise but more-efficient approximation is to drop the absolute value function:

%$$\bm{s} \approx \sqrt[]{\sum_k (\frac{\sum_i \nabla_{\bm{w}} NN(\bm{X}_i)_k}{I})^2},$$ 
$$\bm{s}_{\textrm{SUM}} \approx \sqrt[]{\sum_k \bigg(\sum_i \nabla_{\bm{w}} NN(\bm{X}_i)_k\bigg)^2},$$ 
which is referred to as the \emph{summed gradient} variant, or \emph{SM-G-SUM}. Note that each output gradient is no longer averaged over all $I$ timesteps of experience, which empirically improves performance, potentially by counteracting the washout effect from summing together potentially opposite-signed gradients. That is, if modifying a weight causes positive and negative output changes in different experiences, the measured sensitivity of that weight would be \emph{washed-out} in the sum of these opposite-signed changes (see the Gradient Washout task in supplemental material section 1 for a toy example); later experiments explore whether such dilution imposes a cost in practice.

A final SM-G approach is to consider gradients of the divergence equation (equation \ref{eq:divergence}) itself, i.e.\ how does perturbing a weight affect the formalism this paper adopts to define safe mutations in the first place. However, a \emph{second-order} approximation must be used, because the gradient of equation \ref{eq:divergence} with respect to the weights is uniformly zero when evaluated at the NN's current weights ($\boldsymbol{\delta}=0$), i.e.\ divergence is naturally a global minimum because the comparison is between two NNs with identical weights. With second-order information evaluated at $\boldsymbol{\delta}=0$, the gradient of the divergence as weights are perturbed along a randomly-chosen perturbation direction $\boldsymbol{\delta_0}$ can be approximated as:

\begin{comment}
\begin{align}
\nabla_{\bm{w}} (\textrm{Divergence}(0+{\boldsymbol{\delta}_0};\bm{w})) &\approx \nabla_{\bm{w}}\textrm{Divergence}(0;\bm{w})+H_{\bm{w}}(\textrm{Divergence}(0;\bm{w}))\boldsymbol{\delta_0} \notag\\
&\approx H_{\bm{w}}(\textrm{Divergence}(0;\bm{w}))\boldsymbol{\delta_0}, \label{eq:sosmog}
\end{align}
\end{comment}

\vspace{-0.15in}
\begin{align}
\nabla_{\bm{w}} (\textrm{Divergence}(0+{\boldsymbol{\delta}_0};\bm{w})) &\approx \nabla_{\bm{w}}\textrm{Divergence}(0;\bm{w})+ \notag\\ & \hspace{0.1in}H_{\bm{w}}(\textrm{Divergence}(0;\bm{w}))\boldsymbol{\delta_0} \notag\\
&\approx H_{\bm{w}}(\textrm{Divergence}(0;\bm{w}))\boldsymbol{\delta_0}, \label{eq:sosmog}
\end{align}
where $H$ is the Hessian of $\textrm{Divergence}$ with respect to $\bm{w}$. While calculating the full Hessian matrix of a large NN would be prohibitively expensive, the calculation of a Hessian-vector product (i.e.\ the final form of equation \ref{eq:sosmog}) imposes only the cost of an additional backwards pass; the insight is that the full Hessian is not necessary because what is important is the curvature in only a single particular random direction in weight space $\boldsymbol{\delta}$. Given this estimate of the gradient in the direction of the mutation $\boldsymbol{\delta}$, per-weight sensitivity for this second-order SM-G (i.e.\ \emph{SM-G-SO}) can then be calculated in a similar way to SM-G-ABS:

\begin{equation*}
%\bm{s}_{\textrm{SO}} = \sqrt[]{\nabla_{\bm{w}} \textrm{Divergence}(\boldsymbol{\delta}+\boldsymbol{\delta}_0)} \\
\bm{s}_{\textrm{SO}} = \sqrt[]{\textrm{abs}(H_{\bm{w}}(\textrm{Divergence}(0;\bm{w}))\boldsymbol{\delta})}, \\
\end{equation*}
and the perturbation $\bm{\delta}$ can similarly be adjusted by dividing by the weight sensitivity vector $\bm{s}_\textrm{SO}$.

\begin{comment}
Note that it is also possible to combine SM-G and SM-R; the idea is that SM-G can adjust weight perturbations according to their sensitivity, and SM-R could operate thereafter to fine-tune the actual effect on the NN to correct approximation errors; preliminary experiments have explored this idea, but it is left to future work to demonstrate its promise. %the experiments explore the effect and necessity of such adjustment.
\end{comment}

\section{Simple Neuroevolution Algorithm}

%KENTODO: You call it a simple "EA" below but elsewhere I think you called it a GA.  Probably worth checking for consistency in terminology.  I think EA is probably the best choice if we have to make one, but we can discuss if there is any uncertainty.
Where not otherwise noted, the experiments in this paper all apply the same underlying NE algorithm, which is based on a simple
steady-state EA (i.e.\ there are no discrete generations and individuals are instead replaced incrementally) with standard tournament selection with size 5. All mutation operators build upon a simple control mutation method based on the successful deep-learning ES of \citet{es},
where the entire parameter vector is perturbed with fixed-variance Gaussian noise. Each NN weight vector composing the initial population is separately randomly initialized with the Xavier initialization rule \citep{glorot:xavier}, which showed promise in preliminary experiments. For simplicity the algorithm does not include
crossover, although there may also be interesting SM-inspired approaches to safe crossover. Evolution proceeds until a solution is evolved or until a fixed budget of evaluations is exhausted. Source code for the NE algorithm and SM operators is available from: \url{https://github.com/uber-common/safemutations}.

%KENTODO: In the paragraph below should we make it clear that controls also have the benefit of a hyperparameter search?  It may not be obvious from this text alone.
The strength of mutations is tuned independently with a grid search for each method on each domain (including the control). For methods besides SM-R such strength corresponds to the variance of the Gaussian weight-vector perturbations, while for SM-R the severity of mutation is varied through adjusting the targeted amount of divergence in equation \ref{eq:divergence}. 
%besides SM-R (including the control); SM-R replaces that hyperparameter with a similarly-intentioned one, i.e.\ the targeted amount of divergence in equation \ref{eq:divergence}. 
Specific mutation strength settings are noted in the supplemental material.
%including mutation-specific settings for each method, which were fit per domain through an initial grid search.

%* Experiments
%   * Domains:
%      * Hard Maze
%         * Simple RL task; can we scale to really deep nets?
%      * Atari Hard Maze? (TODO? This is not done?)
%         * Could show same argument generalizing to convolution domain 
%      * Snake
%         * More complicated markovian RL task
%      * Top-down Maze
%         * Non-markovian task with RNNs
%      * RayCast Maze
%         * Impressive first-person navigation from pixels
%      * Vizdoom (?) -- if enough time
%         * Best approach probably would be NS on ending locations
%      * KEN: Is Mujoco with a GA still on the table?  How’s that going?
%      * KEN: Nondeterministic version of something?
%      * Possible experiment: SMOG more resilient to bad settings
%         * Bad init / bad activation function / etc.

\section{Experiments}

This section explores the scalability of SM techniques by applying them to domains of increasing difficulty, culminating in first-person traversal of a rendered 3D environment from raw pixels.

\vspace{-0.05in}
\subsection{Recurrent Parity Task}\label{sec:rnn}

To highlight a class of NN models for which SM-G might provide natural benefit, this section introduces a simple
recurrent classification task, called \emph{Recurrent Parity}. In this task an NN must classify the parity of a bit-sequence (i.e.\ whether the sequence contains an odd number of ones) 
presented to it sequentially in time. Recurrent networks are known to exhibit vanishing and exploding gradients \citep{hochreiter:lstm}, which 
from a variation point of view would manifest as weights with tiny and massive sensitivities, respectively. 

%KENTODO: Still referring to the "appendix" here - should establish a consistent term when first referenced. hould check throughout paper for "appendix" to make sure we are not missing inconsistency anywhere.
A two-layer RNN network with one input, two recurrent hidden layers of 20 nodes each, and one output is trained to memorize all sixteen 4-bit parities. This network has approximately 1,300 parameters. Twenty independent evolutionary runs are conducted for each method for a maximum of 100,000 evaluations each. Specific hyperparameters for each method are noted in the supplemental material.

Figure \ref{fig:recurrent} shows the fraction of successful runs across evaluations for each run. All SM-G methods evolve solutions significantly faster than the control (Mann-Whitney U-test; $p<0.001$); SM-R's solution time is significantly better than the control only if failed runs of the control are included in the calculation ($p<0.05$).
To support the intuition motivating the SM-G family, i.e.\ that SM-G mutations are safer than control mutations, all solutions \emph{evolved} by each mutation method (20 for both SM-G methods and 17 for Control; 3 Control runs did not solve the task) are subject to 50 \emph{post-hoc} perturbations from each mutation method. The robustness of each individual solution/post-hoc-mutation combination is calculated as the average fraction of performance retained after perturbation. The result is that SM-G methods result in significantly more robust mutations no matter what mutation method generates the solution (Mann-Whitney U-test; $p<0.001$); more details are shown in supplemental material figure 2.

%TODO: possible analysis of mutation distribution for solutions from smog and regular mutation to
%validate safety

%size 20
%20 regularnone 0.05 8 8 49942.875
%20 smogadam 0.001 8 8 26744.875
%20 smogadamso 0.001 8 8 19960.875
%20 smogadamtrpo 0.001 6 6 23300.6666667
%20 smogtrpo 0.005 6 6 39777.3333333

%TODO:
%20 smogadamabs 0.001 5 5 ~30000

%in progress smogtro, smogadamtrpo,smogabs

%size 80
%80 regularnone 0.005 0 8 100000.0   [no solutions across hp sweep...]
%80 smogadam 0.001 7 8 84067.375
%80 smogadamso [no solutions across hp sweep...]
%in progress smogtro, smogadamtrpo, smogabs

%TODO: add SM-G-ABS
%TODO(Joel): Ideally in the key in the figure graphic  the SM-G variants would all be together without one SM-R between them
\begin{figure}
  \centering
  \includegraphics[height=1.8in]{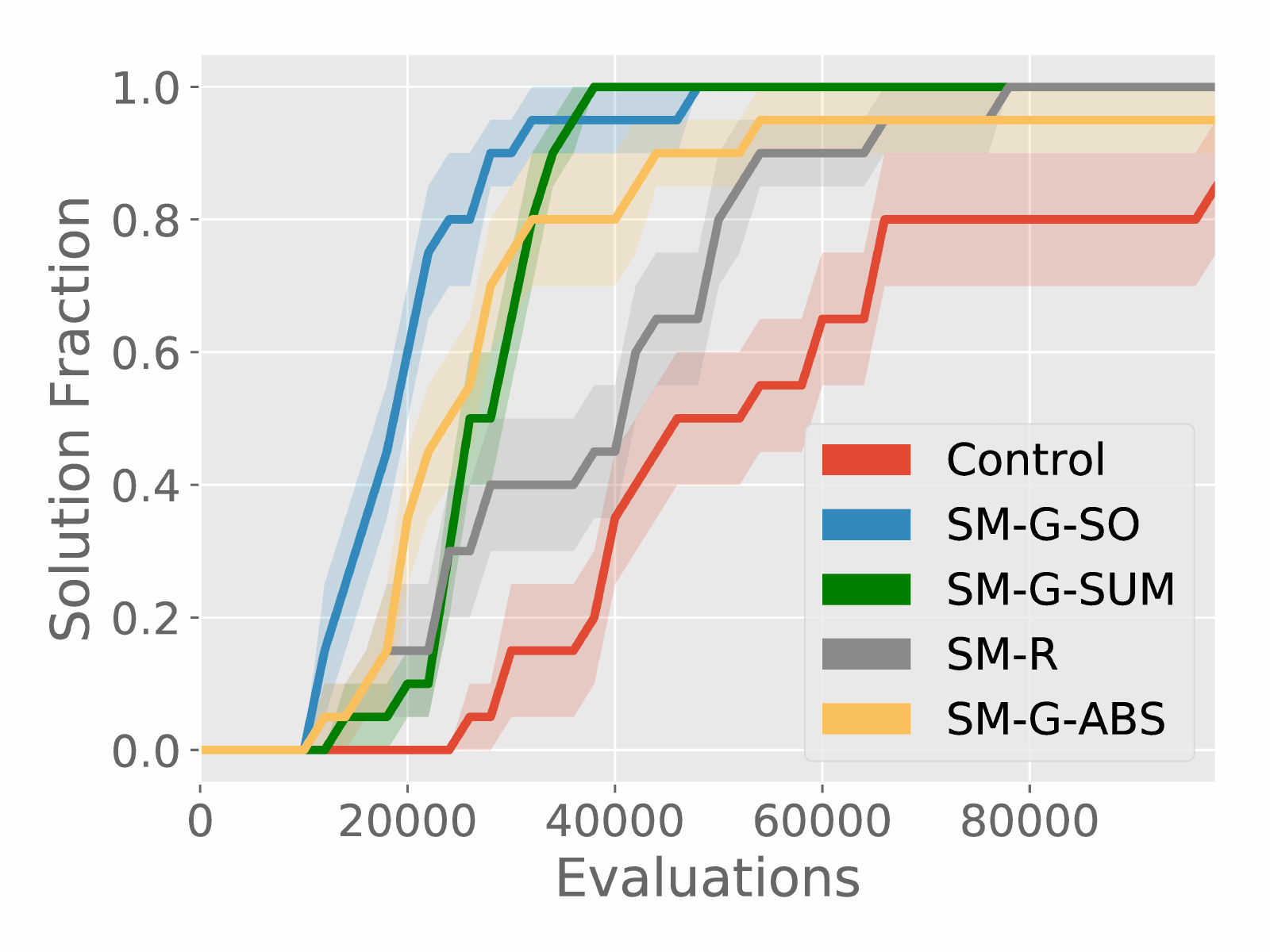}
  %\fbox{\rule[-.5cm]{0cm}{4cm} \rule[-.5cm]{4cm}{0cm}}
	\vspace{-0.25in}
  \caption{\label{fig:recurrent}\textbf{Performance on the Recurrent Parity task across methods.} The plot shows the fraction of solutions evolved by each method across evaluations, for twenty independent runs of each method. Each SM-G approach solves the task significantly faster than the control, highlighting the potential for SM to enhance evolution of recurrent NNs.}
	\vspace{-0.15in}
\end{figure}

\vspace{-0.05in}
\subsection{Breadcrumb Hard Maze}

\begin{figure}
  \centering
  \begin{subfigure}[t]{0.23\textwidth}
  		\centering
         \includegraphics[height=1.1in]{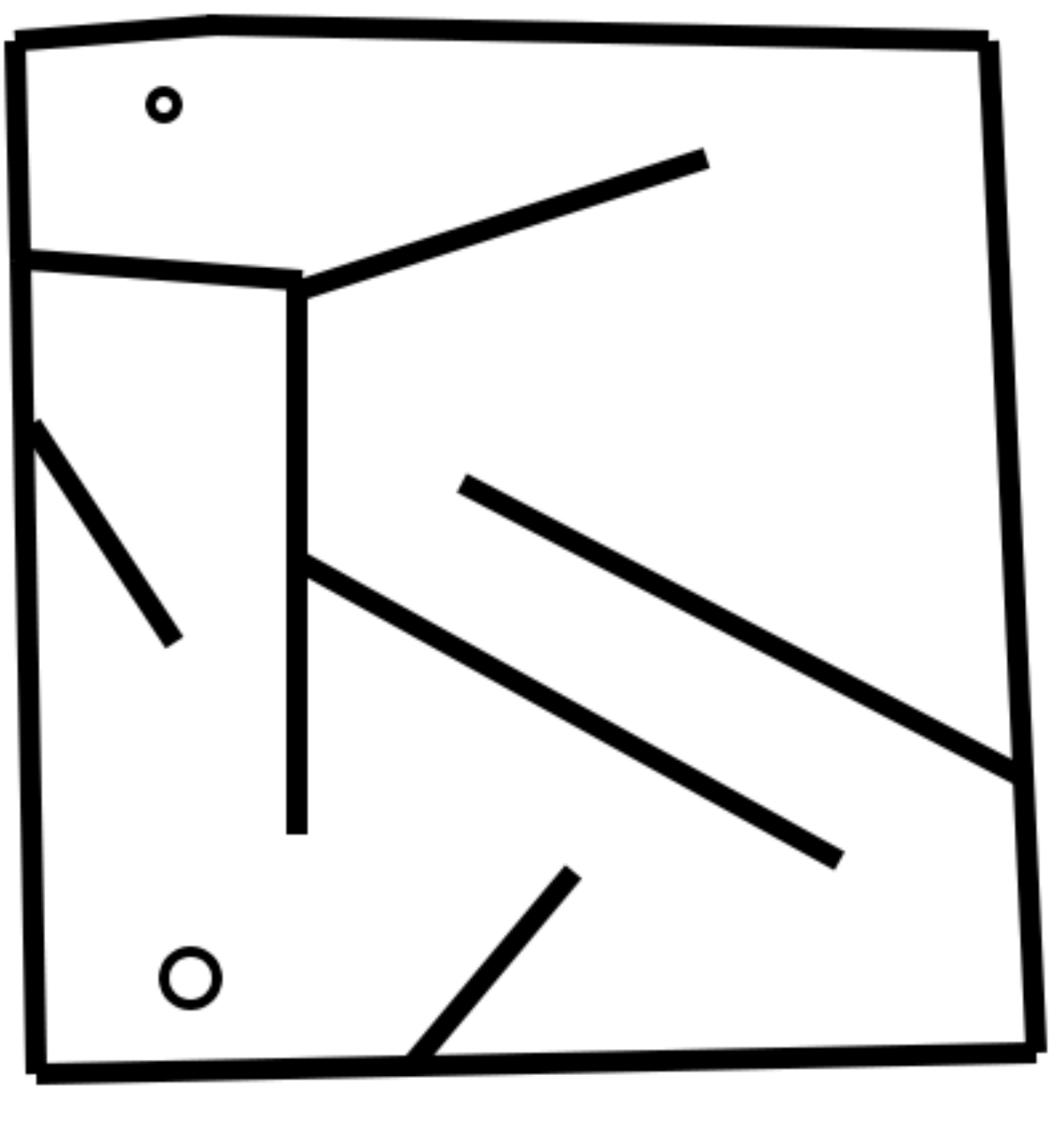}
        \caption{Hard Maze Map}
    \end{subfigure}%
  %\fbox{\rule[-.5cm]{0cm}{4cm} \rule[-.5cm]{4cm}{0cm}}
  \begin{subfigure}[t]{0.23\textwidth}
        \centering
  \includegraphics[height=1.1in]{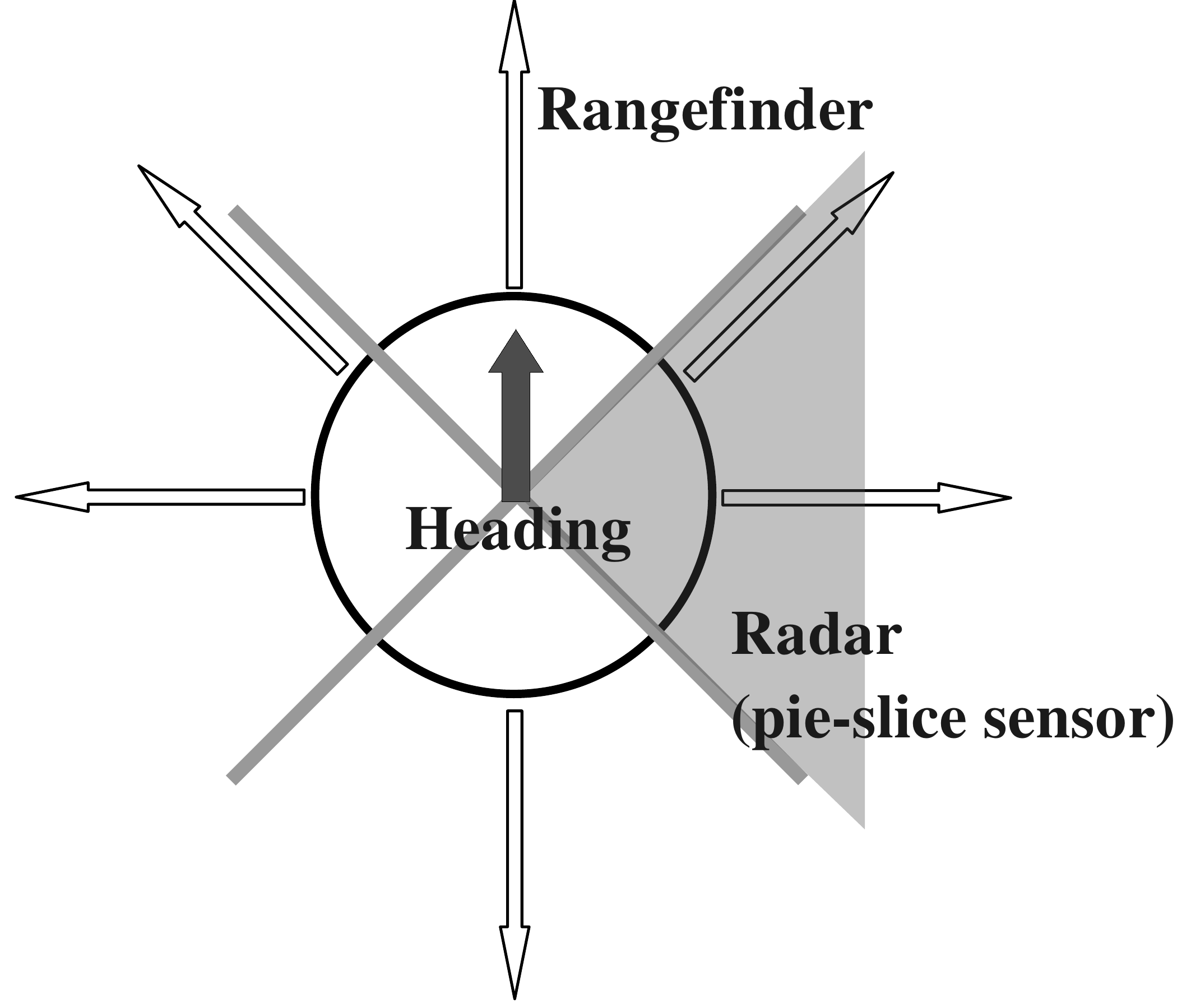}
  \caption{Maze Navigating Robot}
  \end{subfigure}
  \centering
  %\fbox{\rule[-.5cm]{0cm}{4cm} \rule[-.5cm]{4cm}{0cm}}
	\vspace{-0.1in}
	\caption{\label{fig:hardmaze_image}\textbf{The Breadcrumb Hard Maze domain.} The maze's layout is shown in (a), while the maze navigating robot and its sensors is shown in (b). In (a), the large circle represents the robot's starting position and the small circle represents the goal. In (b), each arrow outside of the robot's body is a rangefinder sensor measuring the distance to the closest obstacle in that direction. The robot has four pie-slice sensors that act as a compass, activating when a line from the goal to the center of the robot falls within the pie-slice (i.e.\ irrespective of intervening walls). The solid arrow indicates the robot's heading. Navigating robots are rewarded for ending in locations with low A-star distance to the goal.}
	\vspace{-0.15in}
\end{figure}

The purpose of this experiment is to explore whether SM approaches have promise for evolving deep networks in an RL context. The Breadcrumb Hard Maze (shown in figure \ref{fig:hardmaze_image}) was chosen as a representative low-dimensional continuous control task, and is derived from the Hard Maze benchmark of \citet{lehman:ecj11}. An evolved NN controls a wheeled robot embedded within a maze (figure \ref{fig:hardmaze_image}a), with the objective of navigating to its terminus. The robot receives egocentric sensor information (figure \ref{fig:hardmaze_image}b) and has two effectors that control its velocity and orientation, respectively. 

In its original instantiation, the hard maze was intended to highlight the role of deception in search, and fitness was intended to be a poor compass. Because this work focuses on a different issue (i.e.\ the scalability of evolution to deep networks), the fitness function should instead serve as a reliable measure of progress. Thus, fitness in this breadcrumb version of the Hard Maze domain is rewarded as the negation of the A-star distance to the maze's goal from the robot's location at the end of its evaluation, i.e.\ fitness increases as the navigator progresses further along the solution path (like a breadcrumb trail). Note that a similar domain is applied for similar reasons in \citet{risi:maze}.

Past work applied NEAT to this domain, evolving small and relatively shallow NNs \citep{lehman:ecj11,risi:maze}. In contrast, to explore scaling to deep NNs where mutation is likely to become brittle, the NN applied here consists of 16 feed forward hidden layers of 8 units each, for a total of 1,266 evolvable parameters. The activation function in hidden layers is the SELU \citep{klambauer:selu}, while the output layer has unsigned sigmoid units. The specific hyperparameters for the NE algorithm and mutation operators are listed in the supplemental material.

\subsubsection{Results}

\begin{figure}
  \centering
  \includegraphics[height=2.0in]{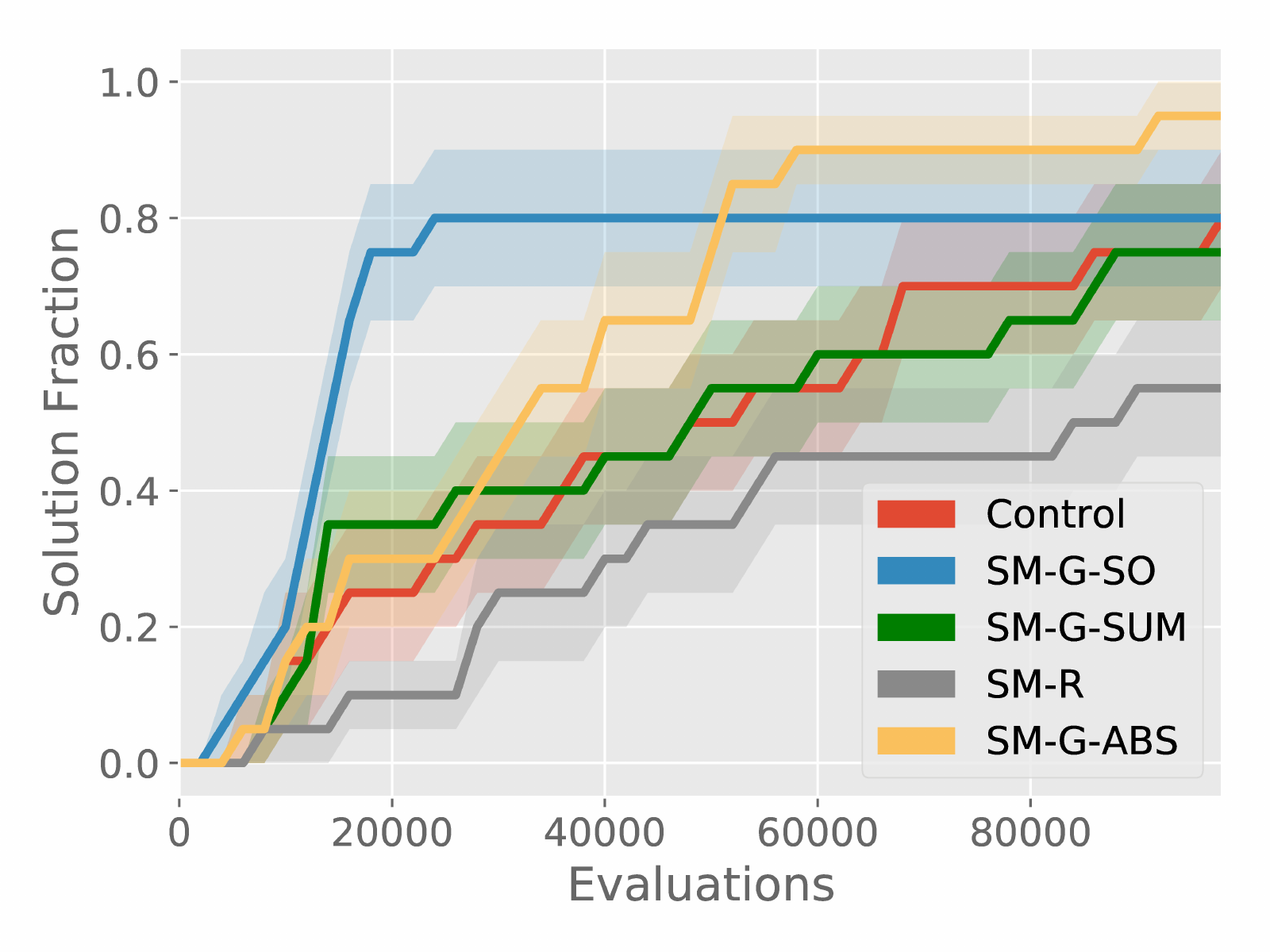}
  %\fbox{\rule[-.5cm]{0cm}{4cm} \rule[-.5cm]{4cm}{0cm}}
	\vspace{-0.2in}
  \caption{\label{fig:hardmaze}\textbf{Performance on the Breadcrumb Hard Maze across methods.} The fraction of solutions evolved by each method is shown over increasing evaluations. SM-G-SO evolves solutions significantly more quickly than the standard mutation control, and only SM-G-ABS evolves solutions in each of its $20$ independent runs. Interestingly, SM-G-SUM's performance mirrors the control, while SM-R under-performs the other methods. The conclusion is that some domains benefit from SM methods that exploit more principled gradient information (SM-G-ABS and SM-G-SO).}
	
\end{figure}

Figure \ref{fig:hardmaze} shows results across different
mutation approaches. SM-G-SO evolves solutions significantly more
quickly than the control (Mann-Whitney U-test; $p<0.005$). The only method
to solve the task in all runs was SM-G-ABS, and although the difference in number of solutions did not
differ significantly from the control at the end of evolution (Barnard's exact test; $p>0.05$), at 60,000 evaluations
it had evolved significantly more solutions than the control and SM-R (Barnard's exact test; $p<0.05$). SM-R performs poorly
in this domain, suggesting that nuanced gradient information may often provided greater benefit to crafting safe variation;
similarly, SM-G-SUM performs no differently from the control, suggesting the gradient washout effect may affect this domain. A video in the supplemental material highlights the qualitative benefits of the SM approaches when mutating solutions.
%\footnote{\url{https://goo.gl/BrKDTj}}.
%\footnote{\url{https://www.youtube.com/playlist?list=PLxWSC7x4MS2dWFvG78gi8EUXK2s02_HEB}}.

%TODO probably split into two parallel figures if possible.

%Hyperparameter search across methods:
%For 16-layer selu nets in hard maze

%FROM maze_hp_second folder [which contains regular runs and confusingly, absgrad]
%regularnone 0.05 8 10 65056.9 [43771.0, 100000.0, 62649.0, 42267.0, 28633.0, 86003.0, 35011.0, 84361.0, 100000.0, 67874.0]
%abs-smog still running...
%abs-smog 0.005 4 4 11169.0 [9186.0, 7940.0, 10037.0, 17513.0]

%FROM maze_hp_secondorder folder [which contains so-smog results]
%RERUN hp because of change
%so-smog 0.05 10 10 22825.9 [8990.0, 22613.0, 17928.0, 11692.0, 7256.0, 70117.0, 6181.0, 66914.0, 9199.0, 7369.0]

%FROM maze_hp folder [note not all complete yet...]
%avg-smog 0.01 8 9 34852.2222222 [36989.0, 31770.0, 12381.0, 23516.0, 100000.0, 19706.0, 21889.0, 45499.0, 21920.0]
%avg-smog+ir 0.01 8 8 21291.875 [21084.0, 14650.0, 15876.0, 58181.0, 11515.0, 16346.0, 12503.0, 20180.0]
%smir 0.005 8 8 26307.125 [16815.0, 65871.0, 33397.0, 4140.0, 13280.0, 40877.0, 11249.0, 24828.0]

\vspace{-0.1in}
\subsubsection{Large-scale NN Experiment}

% 64 layer net with 125 units / layer
% 101 layer net with 48 units / layer

Additional experiments in this domain explore the ability of SM-G methods to
evolve NNs with up to a hundred layers or a million parameters; before now neuroevolution of NNs with over 10 layers has little precedent. 
Three network architectures inspired by wide residual networks \citep{zagoruyko:wide} were tested, consisting of 32, 64, and 101 Tanh layers, with residual skip-ahead connections every four layers. 
The 32 and 64-layer models are designed to explore parameter-size scalability, and have 125 units in each hidden layer, resulting in models with approximately half a million and a million parameters, respectively. The 101-layer model is designed instead to explore scaling NE to extreme depth; each layer contains fewer units (48 vs.\ 125) than the 32 and 64-layer models, resulting in fewer total parameters (approximately 200,000) despite the NN's increased depth.
While the previous experiment shows that such capacity (a million parameters or 100 layers) is unnecessary to solve this task, success nonetheless highlights the potential for NE to scale to large models. Note that hyperparameters are fit using grid search on the 32-layer model, and are then also applied to the 64 and 101-layer models. 

Figure \ref{fig:bigmaze} shows results across different mutation methods for each of the three architectures, with $20$ independent runs for each combination of model and method. In each case, SM-G-SUM evolves significantly more solutions than either SM-G-SO or the control ($p<0.05$; Barnard's exact test). The conclusion is that SM-G shows potential for effectively evolving large-scale NNs. Note that more detailed training plots are included in the supplemental material. 

\begin{figure} %[h]
	\vspace{-0.1in}
  \includegraphics[height=1.85in]{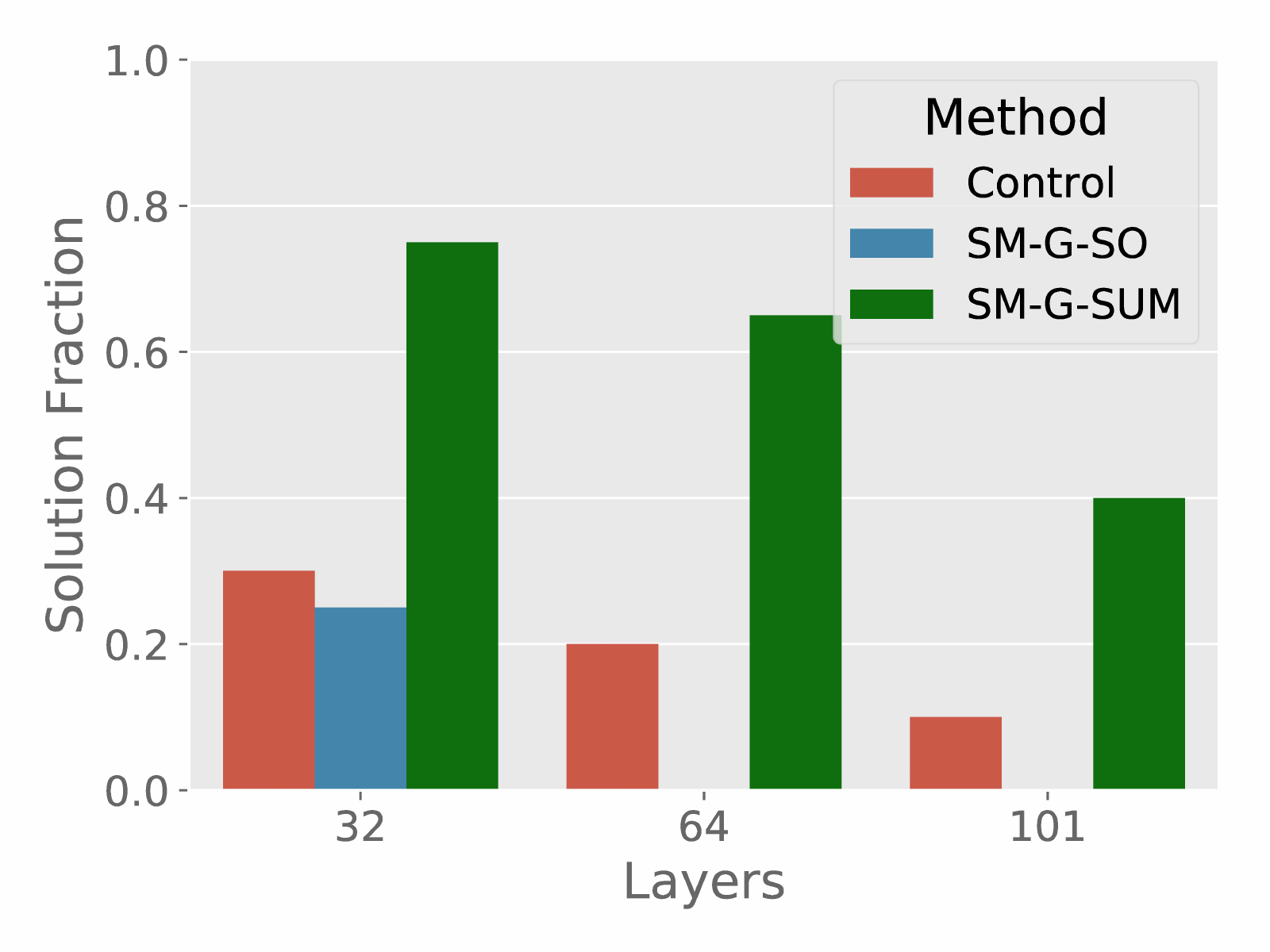}
	\vspace{-0.2in}
  \caption{\label{fig:bigmaze}\textbf{Performance on the large-scale NN task across methods.} The fraction of solutions successfully evolved by each method over 20 runs is shown. Although performance degrades with increasing layers for all methods, SM-G-SUM evolves significantly more solutions than the standard mutation control and SM-G-SO in each of the 32-layer, 64-layer, and 101-layer models. The conclusion is that SM-G can help to unlock the potential of NNs with up to a million parameters or a hundred layers.}
	\vspace{-0.15in}
\end{figure}
\vspace{-0.1in}

\subsection{Topdown Maze}

The Topdown Maze domain is designed to explore whether safe mutation can accelerate the evolution of deep 
recurrent convolutional NNs that learn from raw pixels. The motivation is that this is a powerful architecture 
that enables learning abstract representations without feature engineering as well as integrating information over time; similar combinations of recurrence and convolution have shown considerable 
promise in deep RL \citep{mirowski:recurrent}. 

\begin{figure}
  \centering
   
  \begin{subfigure}[t]{0.20\textwidth}
  		\centering
	\includegraphics[height=0.8in]{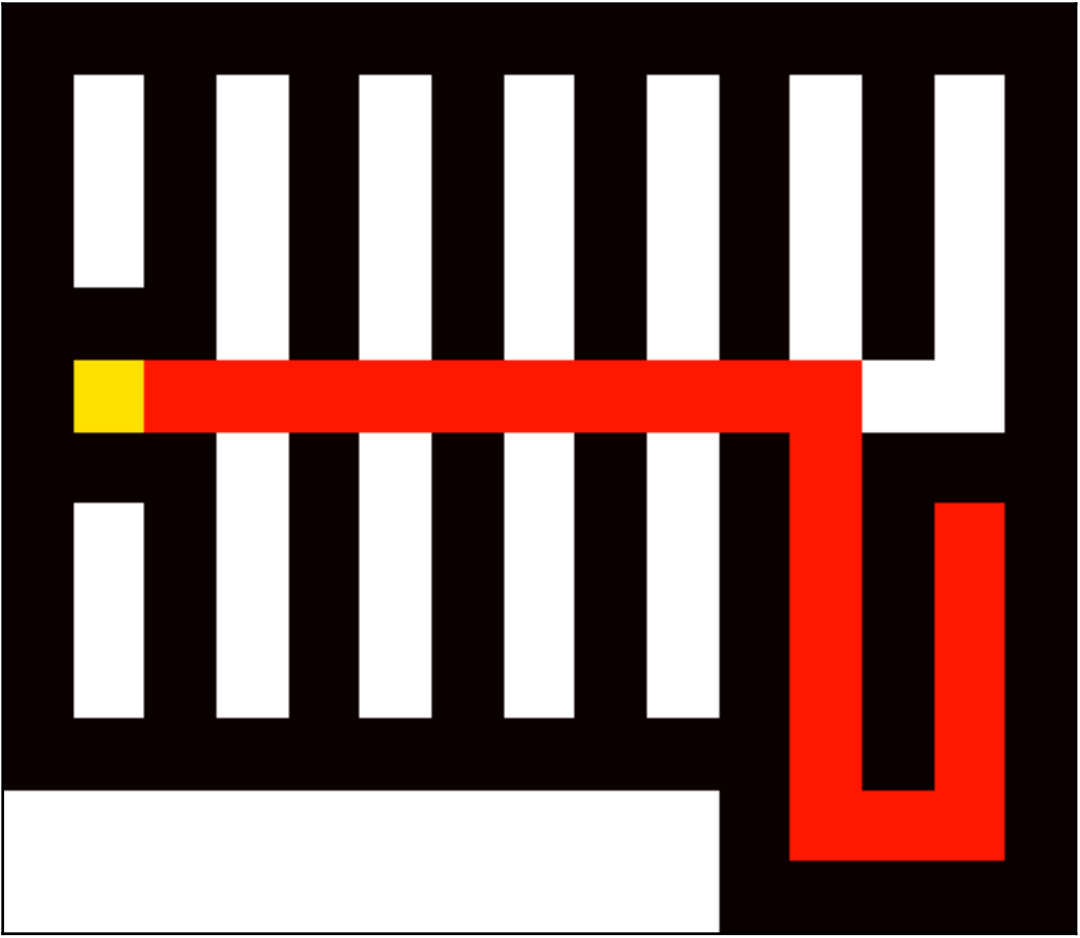}
	 \caption{Map}
  \end{subfigure}%
  \begin{subfigure}[t]{0.15\textwidth}
  		\centering
        \fbox{ \includegraphics[height=0.7in]{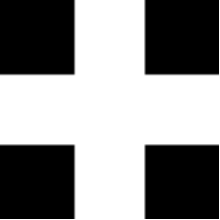}}
        \caption{Intersection}
    \end{subfigure}%
  %\fbox{\rule[-.5cm]{0cm}{4cm} \rule[-.5cm]{4cm}{0cm}}
  \begin{subfigure}[t]{0.15\textwidth}
        \centering
  \fbox{\includegraphics[height=0.7in]{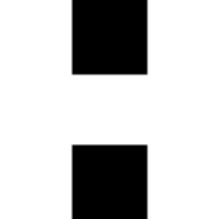}}
  \caption{Inbetween}
  \end{subfigure}
  %\begin{subfigure}[t]{0.15\textwidth}
  %      \centering
  %\fbox{\includegraphics[height=0.7in]{figures/topdown_frame26}}
  %\caption{Near maze end}
  %\end{subfigure}
	\vspace{-0.1in}
  %\fbox{\rule[-.5cm]{0cm}{4cm} \rule[-.5cm]{4cm}{0cm}}
	\caption{\label{fig:topdown_illustrate}\textbf{Topdown Maze domain.} The (a) 2D grid-world maze is shown in which the agent is embedded. A black square indicates a wall and the red path indicates the target trajectory of the agent. One fitness point is awarded for each square along the trajectory the agent touches. Note that the red trajectory is not visible to the agent. Because the agent views only a $3\times3$ block  window around its immediate location, many (b) intersections and (c) positions between intersections are conflated. Successful completion of the maze thus requires integrating information over time by making use of recurrence. Note that each block is rendered as a $21\times21$ square of the NN's input image.} 
	
\end{figure}

In this domain, the agent receives a visual $64$x$64$ input containing a local view of a maze (i.e.\ it cannot see the whole maze at once) as a grayscale image, and has discrete actions that navigate one block in each of the four cardinal directions.
Because the maze (figure \ref{fig:topdown_illustrate}a) has many identical intersections, which the agent cannot 
distinguish by local visual information alone (figures \ref{fig:topdown_illustrate}b and c), solving the task requires use of recurrent memory.

These experiments focus on comparing the more computationally-scalable variants of SM-G (i.e.\ SM-G-SUM and SM-G-SO) to the control mutation method, because of the complexity and size of the RNN. Note that because this NN is recurrent, backpropagation through time
is used for the SM-G approaches when calculating weight sensitivity, i.e.\ weight sensitivity in SM-G is 
informed by the cascading effects of signals over time.

\begin{comment}
\begin{figure}[h]
  \centering
  \includegraphics[height=1.4in]{figures/topdown_render}
  %\fbox{\rule[-.5cm]{0cm}{4cm} \rule[-.5cm]{4cm}{0cm}}
	\vspace{-0.15in}
	\caption{\label{fig:topdown_illustrate}\textbf{Topdown Maze domain.} The 2D grid-world maze is shown in which the agent is embedded. A black square indicates a wall and the red path indicates the target trajectory of the agent. The agent receives one fitness point for each square along the trajectory it touches. Note that the red trajectory is not visible to the agent.}
\end{figure}
\end{comment}

%Hyperparameter search 
%regularnone 0.1 8 11 54935.5454545
%smogadam 0.01 10 10 2077.7
%smogadamso 0.01 9 9 3118.88888889 

\subsubsection{Experimental Settings}

The agent receives as input a $64\times64$ grayscale image and has at most $40$ time-steps to navigate the environment.
The NN has a deep convolutional core, with two layers of $5\times5$ convolution with stride 2, to reduce dimensionality, followed by $12$ layers of $3\times3$ convolution with stride 1. All convolutional layers have 12 feature maps and SELU activation. This pathway feeds into an average pooling layer that leads into a two-layer LSTM \citep{hochreiter:lstm} recurrent network with 20 units each; the signal then feeds into an output layer with sigmoid units. The NN has in total 25,805 evolvable parameters and 17 trainable layers. EA and mutation hyperparameters are provided in the supplemental material.

\subsubsection{Results}

\begin{figure}
  \centering
  \includegraphics[height=1.5in]{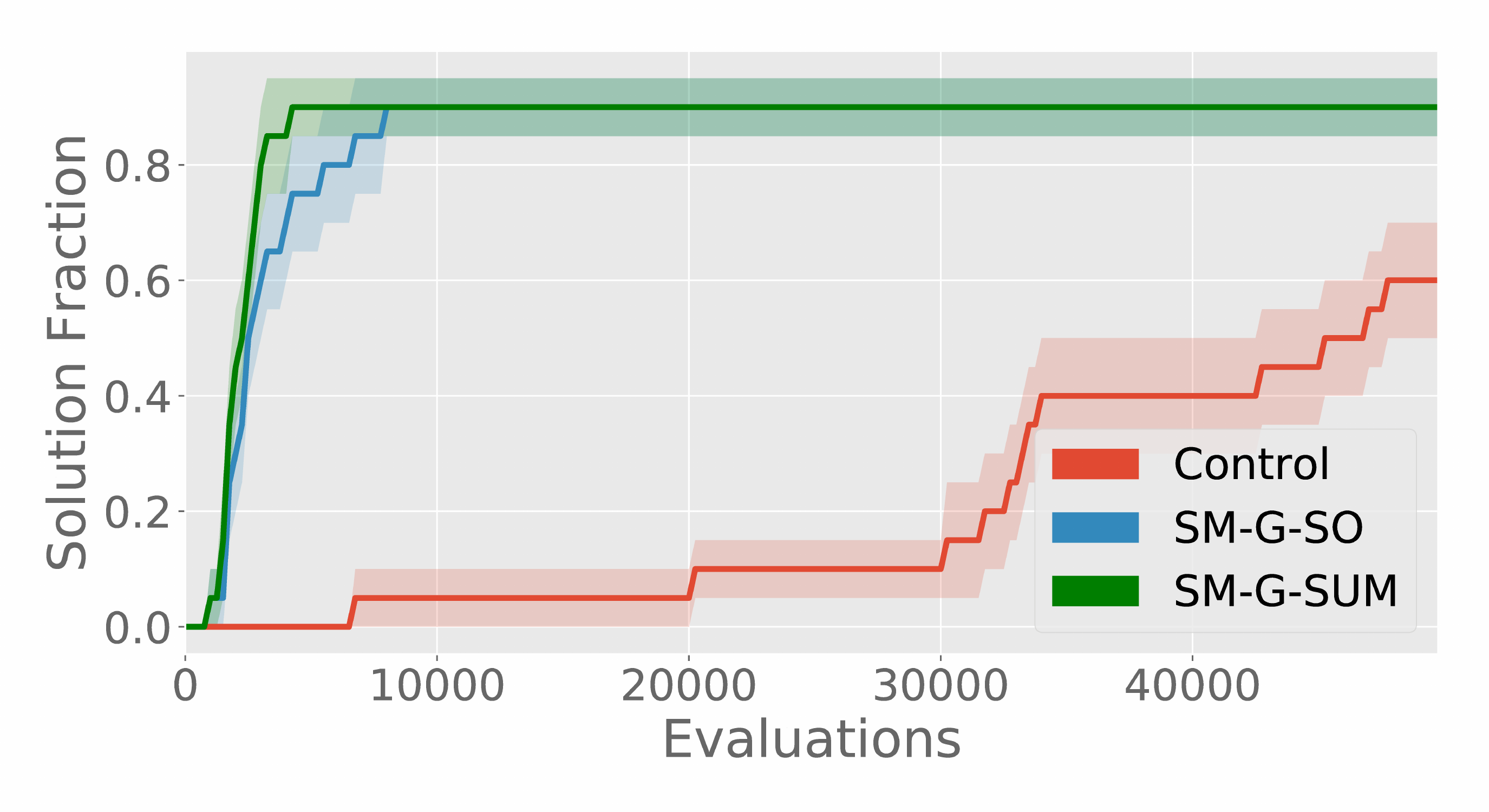} %was 2.1
  %\fbox{\rule[-.5cm]{0cm}{4cm} \rule[-.5cm]{4cm}{0cm}}
	\vspace{-0.2in}
  \caption{\label{fig:topdown}\textbf{Performance on the Topdown Maze across methods.} The fraction of successful independent runs (from the $20$ conducted for each method) is shown across SM-G methods and the control mutation method. SM-G-SUM and SM-G-SO solve the task consistently and in relatively few evaluations when compared to the control.
	}
	\vspace{-0.15in}
\end{figure}

Figure \ref{fig:topdown} shows the results in this domain. Both SM-G-SUM and SM-G-SO consistently evolve solutions, and even when the control is successful it requires significantly more evaluations for it to discover a solution (Mann-Whitney U-test; $p<0.001$). Complementing the results of the simple RNN classification task in section \ref{sec:rnn}, the conclusion from this experiment is that the tested SM-G approaches can accelerate evolution of successful memory-informed navigation behaviors. 

\subsection{First-person 3D Maze}

%TODO: HP search and runs 
%regularnone 0.005 2 3 43515.0
%smogadamso 0.01 3 3 5251.0
%smogadam 0.01 2 2 9340.0

\begin{figure}
  \centering
  \includegraphics[height=1.2in]{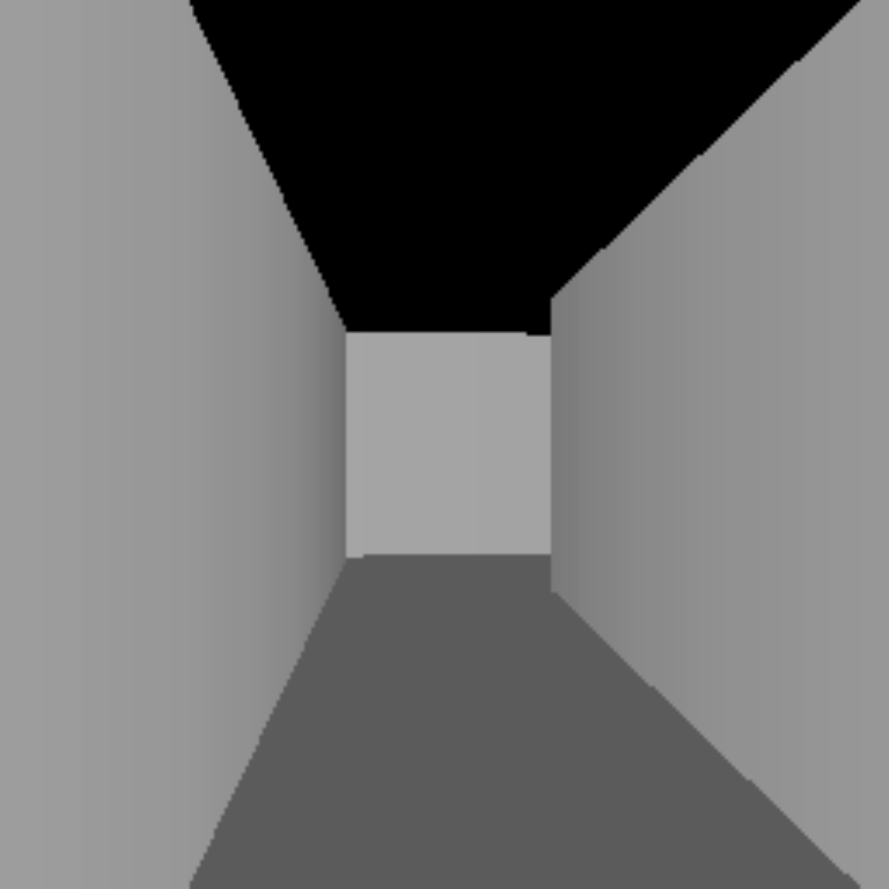}
  %\fbox{\rule[-.5cm]{0cm}{4cm} \rule[-.5cm]{4cm}{0cm}}
  \vspace{-0.1in}
  \caption{\label{fig:raycast_illustrate}\textbf{First-person 3D Maze Domain.} An agent traverses a 3D first-person environment and is rewarded the further it progresses along the correct path.}
	\vspace{-0.05in}
\end{figure}

The First-person 3D Maze is a challenge domain in which a NN learns to navigate an environment from first-person 3D-rendered images (figure \ref{fig:raycast_illustrate}). The maze has the same
layout as the Topdown Maze. However, navigation is egocentric and continuous in space and heading, i.e.\ the agent does not advance block-wise in cardinal directions, but its four discrete actions incrementally turn the agent left or right, or advance or reverse. The agent is given $400$ frames to navigate the maze. Note that this domain builds upon the RayCast Maze environment of the PyGame learning environment \citep{tasfi:PLE}.

%TODO: Rendering of top-down view of the maze (which is slightly different from top-down maze)
%\vspace{-0.05in}

\subsubsection{Experimental Settings}

%KENTODO: again "appendix" comes up below
Input to the NN is a grayscaled $64\times64$ image.
The NN has an architecture nearly identical to that of the ANN of the Topdown Maze, i.e.\ a deep convolutional core (but with 8 instead of 12 layers) feeding into a two-layer LSTM stack (each composed of 20 units), which connects to an output layer with sigmoid units. There are 20,573 parameters in total. The NN executes the same action $4$ frames in a row to reduce the computational expense of the domain, which is bottlenecked by forward propagation of the RNN (for the control mutation method) and a combination of forward and backward RNN propagation (for SM-G approaches). The EA settings are the same as in the Topdown maze, but hyperparameters for each mutation method are fit for this domain separately and are described in the supplemental material.

\vspace{-0.05in}
\subsubsection{Results}

\begin{figure}
  \centering
  \includegraphics[height=1.5in]{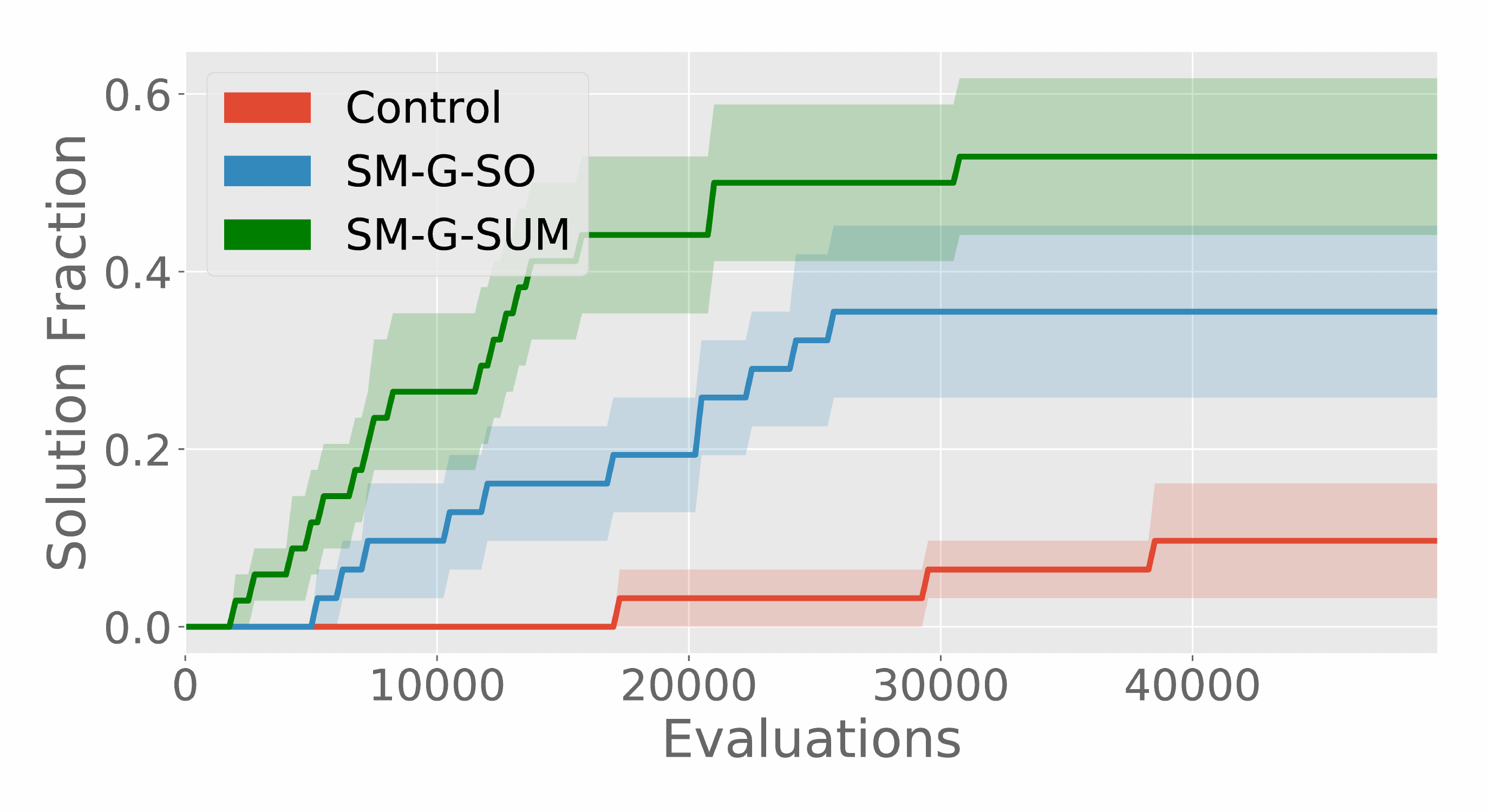}
  %\fbox{\rule[-.5cm]{0cm}{4cm} \rule[-.5cm]{4cm}{0cm}}
  \vspace{-0.25in}
  \caption{\label{fig:rcmaze}\textbf{Performance on the First-person 3D Maze across methods.} The fraction of successful independent runs (from the 30 conducted for each method) is shown across SM-G methods and the control mutation method. SM-G-SUM and SM-G-SO both solve the task significantly more frequently than does the control. }
\vspace{-0.2in}
\end{figure}

%KENTODO: I think the video you mention below is the second one we promised - just be careful that our supplemental materials indeed include precisely the videos we promise it will
Figure \ref{fig:rcmaze} shows the results in this domain. As in the Topdown Maze domain, both SM-G-SUM and SM-G-SO solve the task significantly more often than does the control ($p<0.05$; Barnard's exact test). The conclusion is that SM-G methods can help scale NE to learn adaptive behaviors from raw pixel information using modern deep learning NN architectures. A video of a solution from SM-G-SUM is included in the supplemental material.

%\subsection{VizDoom}

%use same hp and setup as raycast maze
%regularnone 0.005 2 3 43515.0
%smogadamso 0.01 3 3 5251.0
%smogadam 0.01 2 2 9340.0

%HERE
%\vspace{-0.1in}
\section{Discussion and Conclusion}

The results across a variety of domains highlight the general promise of safe mutation operators, in particular when guided
by gradient information, which enables evolving deeper and larger recurrent networks. Importantly,
because SM methods generate variation without taking performance into account, they can easily apply to
artificial life or open-ended evolution domains, where well-defined reward signals may not exist; SM techniques are also agnostic to the stationarity or uni-modality of reward, and can thus be easily applied to
divergent search \citep{lehman:ecj11,mouret:arxiv15} and coevolution \citep{popovici:hnc12}. 
Additionally, SM-G may well-complement differentiable \emph{indirect encodings} where mutations can have outsize impact; for example, in HyperNEAT \citep{stanley:alife09}, CPPN mutations could be constrained such that they limit the change of connectivity in the substrate (i.e.\ target network) or the substrate's output. This safety measure would not preclude systematic changes in weight, only that those systematic changes proceed \emph{slowly}. While both SM-G and SM-R offer alternative routes to safety, it appears from the initial results in this paper that SM-G (and variants thereof) is likely the more robust approach overall.

%-opening up to DL in general
Another implication of safe mutation is the further opening of NE to deep learning in general.  Results like the recent revelation from \citet{es} that an evolution strategy (ES) can rival more conventional deep RL approaches in domains requiring large or deep networks such as humanoid locomotion and Atari have begun to highlight the role evolution can potentially play in deep learning; interestingly, the ES of \citet{es} itself has an inherent drive towards a form of safe mutations \citep{lehman:arxiv17fd}, highlighting the general importance of such mutational safety for deep NE. 
%JOELNOTE: The sentence above is where I linked to ES and safety
Some capabilities, such as indirect encoding or searching for architecture as in classic algorithms like NEAT \citep{stanley:ec02}, are naturally suited to NE and offer real potential benefits to NN optimization in general.  Indeed, combinations of NE with SGD to discover better neural architectures are already appearing \citep{liu:arxiv17,miikkulainen:arxiv17}.  The availability of a safe mutation operator helps to further ease this ongoing transition within the field to much larger and state-of-the-art network architectures.  A wide range of possible EAs can
%KENTODO: Maybe take out "(beyond only ES)" below - even if not intended, it makes it sounds like we are implying that ES is the only algorithm that benefited from SM so far, which is really confusing given that ES is not the algorithm used in this paper.
benefit, thereby opening the field anew to exploring novel algorithms and ideas.   

%-big computation
In principle the increasing availability of parallel computation should benefit NE.  After all, evolution is naturally
parallel and as processor costs go down, parallelization becomes more affordable.  However, if the vast majority of mutations
in large or deep NNs are destructive, then the windfall of massive parallelism is severely clipped.  In this way, SM-G can play
an important role in realizing the potential benefits of big computation for NE in a similar way that innovations such as ReLU activation \citep{glorot:icais2011} (among many others) in deep learning have allowed researchers to capitalize on the increasing power of GPUs in passing gradients through deep NNs.

%-stealing the gradient
In fact, one lingering disadvantage in NE compared to the rest of deep learning has been the inability to capitalize on explicit gradient information when it is available.  Of course, the quality of the gradient obtained can vary -- in reinforcement learning for example it is generally only an indirect proxy for the optimal path towards higher performance -- which is why sometimes NE can rival methods powered by SGD \citep{es,such:arxiv17}, but in general it is a useful guidepost heretofore unavailable in NE.  That SM-G can now capitalize on gradient information for the purpose of safer mutation is an interesting melding of concepts that previously seemed isolated in their separate paradigms.  SM-G is still distinct from the reward-based gradients in deep learning in that it is computing a gradient entirely without reference to how it is expected to impact reward, but it nevertheless capitalizes on the availability of gradient computation to make informed decisions.  In some contexts, such as in reinforcement learning where the gradient may even be misleading, SM-G may offer a principled alternative -- while we sometimes may not have sufficient information to know \emph{where} to move, we can still explore different directions in parallel as safely as possible.  That we can do so without the need for further rollouts (as required by e.g.\ \citet{peng:gpo}) is a further appeal of SM-G.

%JOELNOTE: Added this to try and hit on the idea of targeted mutations
Furthermore, this work opens up future directions for understanding, enhancing, and extending the safe mutation operators introduced here. For example, it is unclear what domain properties predict which SM or SR method will be most effective. Additionally, similarly-motivated safe \emph{crossover} methods could be developed, suggesting there may exist other creative and powerful techniques for exploiting NN structure and gradient information to improve evolutionary variation. 
%An additional interesting future direction is to produce specific, random behavioral variation directly via gradients. This idea seems at first similar to policy gradients, but is different because the effects of changes to state-action mappings for one state also effect other states when compressed state representations like DNNs are used, meaning such a method produces consistent, temporally extended exploration, which has recently been shown to be promising in RL \citep{TODO}. Preliminary experiments explored this idea but without success;it is left to future work to further investigate its promise.
Highlighting another interesting future research direction, preliminary experiments explored a version of SM-G that exploited supervised learning to program a NN to take \emph{specific} altered actions in response to particular states (similar in spirit to a random version of policy gradients for exploration); such initial experiments were not successful, but the idea remains intriguing and further investigation of its potential seems merited.

%JOELNOTE: took out the note about size of NNs, given that larger NNs have been evolved (felipe's has millions of parameters)
Finally, networks of the depth evolved with SM-G in this paper have never been evolved before with NE, and those with similar amounts of parameters have rarely been evolved \citep{such:arxiv17}; in short, scaling in this way might never have been expected to work.  In effect, SM-G has dramatically broadened the applicability of a simple, raw EA across a broad range of domains.  The extent to which these implications extend to more sophisticated NE algorithms is a subject for future investigation.  At minimum, we hope the result that safe mutation can work will inspire a renewed interest in scaling NE to more challenging domains, and reinvigorate initiatives to invent new algorithms and enhance existing ones, now cushioned by the promise of an inexpensive, safer exploration operator.

%* Discussion
%   * Combining with open-ended evolution
%   * Combining with indirect encoding
%      * For HyperNEAT could make sure child CPPN doesn’t diverge too far from parent CPPN; or could go further and examine the substrates themselves
%         * KEN: When you say child CPPN doesn’t diverge too far from parent, do you mean the substrate?  And by “go further” do you mean to the behavior itself?  When it comes to the CPPN itself, I don’t understand how you can make sure it doesn’t diverge too far when it’s the genotype?  Maybe I’m missing something.
%   * Trade-off between incremental change and “hopeful monsters”?
%   * Highlight that networks of this depth/size have never before been run with NE and perhaps were never expected to

% NOTE: Interestingly, some things may work diffrently *with* smog than *without* it; e.g. in concurrent work 
% reg evolution seems to work poorly with LSTMs while SMOG seems improved by it; similarly in initial
% experiments, *WIDE* resnets worked well with SMOG but poorly with regular mutation.

% NOTE: Could be combined with co-evolution etc.

% NOTE: In discussion admit that SMOG assumes that it is often more effective to measure divergence in ouptuts
% rather than in parameter space; but SMOG may lose effectiveness if outputs vary hugely in scale, or if there
% are outputs that themselves are hugely more performance-sensitive than others, etc.

\vspace{-0.1in}
\section*{Acknowledgements} 
We thank the members of Uber AI Labs, in particular Thomas Miconi and Xingwen Zhang for helpful discussions. 
We also thank Justin Pinkul, Mike Deats, Cody Yancey, and the entire OpusStack Team inside Uber for providing resources and technical support.

%\section*{References}
\vspace{-0.05in}
\bibliography{smog,nn,ucf}
\bibliographystyle{apalike}

\newcommand{\beginsupplement}{%
	        \setcounter{table}{0}
		        \renewcommand{\thetable}{S\arabic{table}}%
			        \setcounter{figure}{0}
				        \renewcommand{\thefigure}{S\arabic{figure}}%
					     }

\beginsupplement
\pagebreak
\setcounter{section}{0}

	\begin{center}
		\textbf{\LARGE Supplemental Material}
	\end{center}

	%\section*{Supplemental Material}

\section{Simple Poorly-Conditioned Model}\label{sec:simple}

%TODO: Statistically significant number of runs + analysis

To gain a clearer intuition of the benefits and costs of SM-R and SM-G variants, it is useful to introduce a 
toy model constructed purposefully with parameters that significantly vary in their sensitivity:
\begin{equation}
\begin{split}
\bm{y}_0 &= 100\bm{w}_0\bm{x}_0 \\
\bm{y}_1 &= 0.1\bm{w}_1\bm{x}_1,
\end{split}
\end{equation}
where $\bm{y}$ are the model's outputs as a function of the inputs $\bm{x}$ and the current weights $\bm{w}$.
Notice that $\bm{w}_0$ will have 1,000 times the effect on $\bm{y}_0$ than $\bm{w}_1$ has on $\bm{y}_1$. If the scale of expected outputs
for $\bm{y}_0$ and $\bm{y}_1$ is similar, then an uninformed mutation operator would have difficulty generating variation that 
equally respects effects on $\bm{y}_0$ and $\bm{y}_1$; likely the effect of mutation on the NN will be dominated by the scale of $\boldsymbol{\delta}_0$.

Now, consider three variations of the task. In the first, called the \emph{Easy} task, there is a single input-output pair to memorize, which depends only on $\bm{w}_1$:

\begin{equation*}
\textrm{input}_0 = 0.0, 1.0 \quad \textrm{target}_0 = 0.0, 1.0
\end{equation*}

In the second, called the \emph{Medium} task, there is again a single input-output pair to memorize, but solving the task requires tuning $\bm{w}_0$ and $\bm{w}_1$, on which mutations have a substantially different effect:

\begin{equation*}
\textrm{input}_0 = 1.0, 1.0 \quad \textrm{target}_0 = 1.0, 1.0
\end{equation*}

In the last task, called the \emph{Gradient-Washout} task, there are two input-output pairs, designed to highlight a potential failure case of SM-G-SUM:

%\begin{equation*}
\begin{alignat*}{3}
\textrm{input}_0 &= 1.0, 1.0  && \quad \textrm{target}_{0} & &= 1.0, 1.0 \\
\textrm{input}_1 &= -1.0, -1.0 && \quad \textrm{target}_{1} \quad & &= -1.0, -1.0 
\end{alignat*}
%\end{equation*}

The Easy task is designed to highlight situations in which all SM-G and SM-R variants will succeed, the Medium task highlights when SM-G approaches will have advantage over SM-R, and the Gradient-Washout task highlights situations wherein SM-G-ABS and SM-G-SO have advantage over SM-G-SUM. In particular, in the Gradient-Washout task the only relevant parameter is $\bm{w}_0$, but due to opposite-sign inputs, the summed gradient of $\bm{y}_0$ with respect to $\bm{x}_0$ is 0. The more informative average \emph{absolute value} gradient (used by SM-G-ABS) is $100$.

In this experiment, a simple hill-climbing algorithm is applied instead of the NE algorithm described in the previous section. The hill-climber is initialized with small zero-centered noise. Runs last for $2,000$ iterations, wherein an offspring from the current champion is generated, and replaces the champion only if its fitness improves upon that of the champion. 

\begin{figure*}[h]
  \centering
  \begin{subfigure}[t]{0.33\textwidth}
  		\centering
         \includegraphics[height=1.45in]{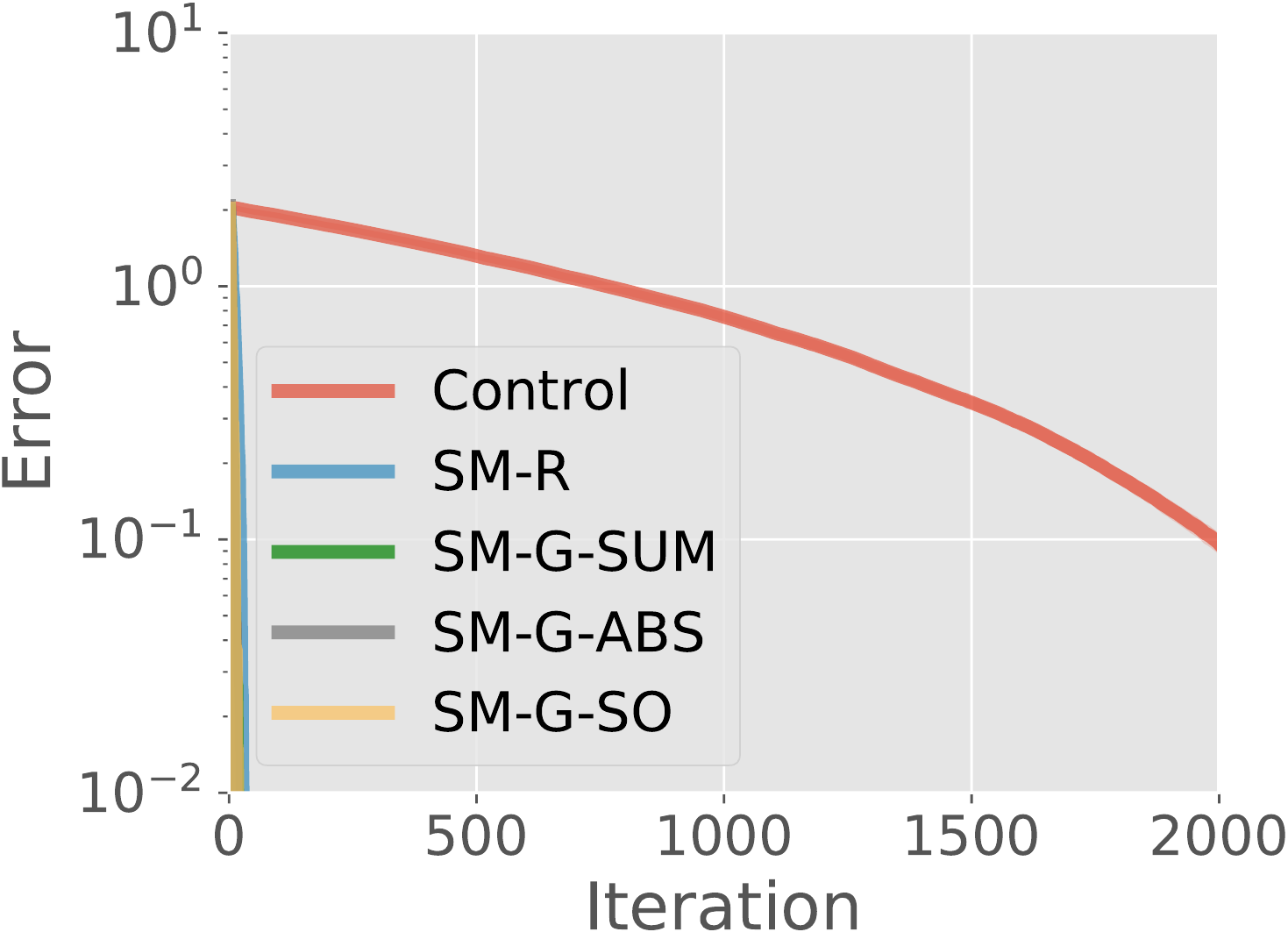}
        \caption{Easy Task}
    \end{subfigure}%
  %\fbox{\rule[-.5cm]{0cm}{4cm} \rule[-.5cm]{4cm}{0cm}}
  \begin{subfigure}[t]{0.33\textwidth}
        \centering
  \includegraphics[height=1.45in]{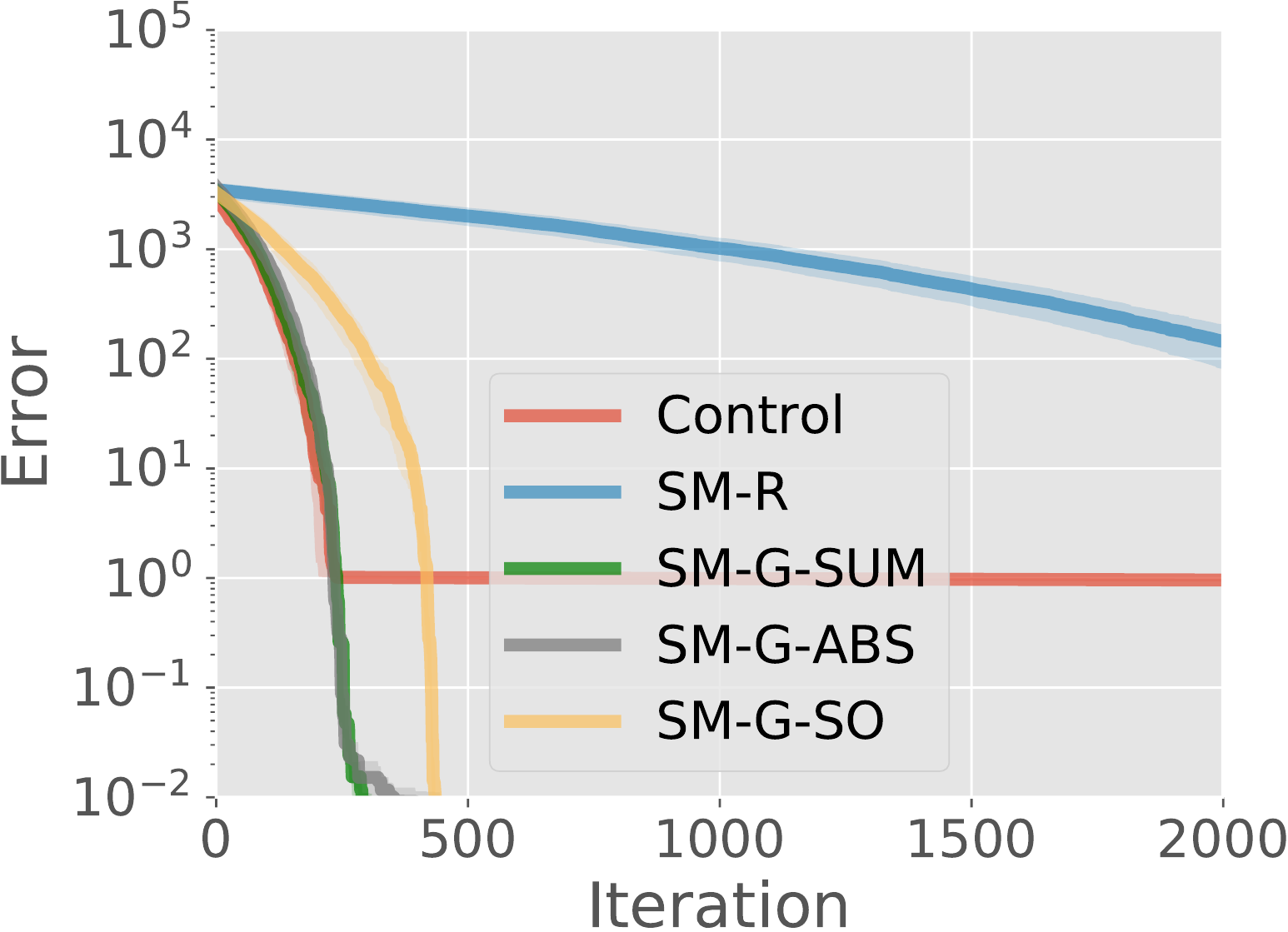}
  \caption{Medium Task}
  \end{subfigure}
  \begin{subfigure}[t]{0.33\textwidth}
        \centering
  \includegraphics[height=1.45in]{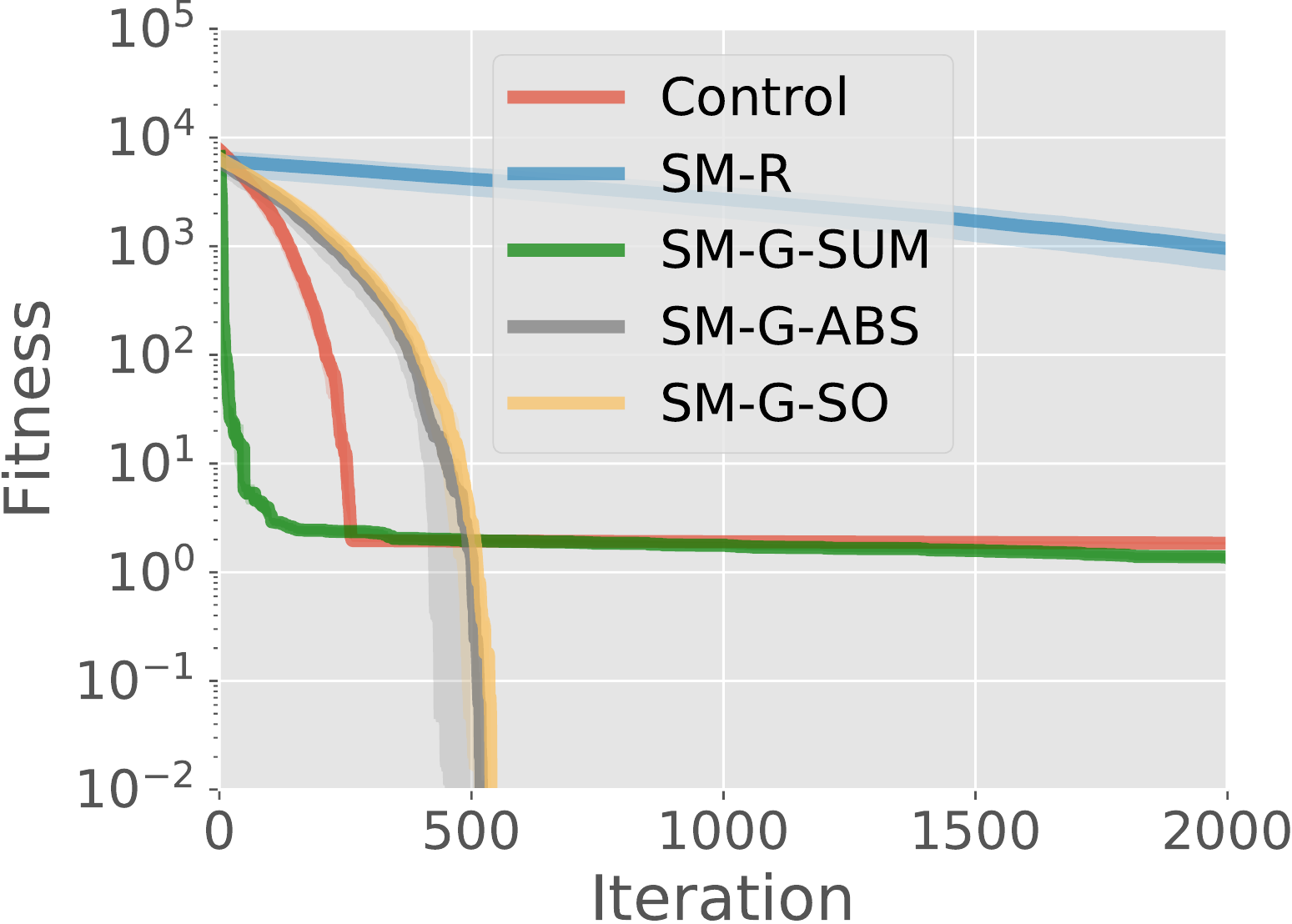}
  \caption{Gradient-Washout Task}
  \end{subfigure}
  \vspace{-0.1in} 
  %\fbox{\rule[-.5cm]{0cm}{4cm} \rule[-.5cm]{4cm}{0cm}}
  \caption{\label{fig:toy}\textbf{Performance of SM-R and SM-G on the Simple Poorly-conditioned Tasks.} All SM-G and SM-R methods perform well on the (a) Easy task, while the (b) Medium task stymies SM-R, and the (c) Gradient-Washout task highlights the benefits of SM-G-ABS and SM-G-SO.}
	
\end{figure*}

Figure \ref{fig:toy} shows the results from $20$ independent runs for each mutation method, i.e.\ control mutation, SM-R, and variations of SM-G. Fitting the
motivation of the experiments, the Easy task is solved effectively by all SM variants (which all outperform the control), the Medium task highlights the benefits of SM methods that take advantage of gradient information, and the Gradient-Washout task highlights the benefits of methods that do not
sum NN outputs over experiences before calculating sensitivity. Note that with a tuned mutation rate, the control can more quickly solve the Easy task, but the point is to highlight that SM-G methods can normalize mutation by their effect on the output of a model, identifying automatically when a parameter is less sensitive (e.g.\ $w_1$ in this task) and can thus safely be mutated more severely. 

%	\section{Additional Figures}

\begin{figure*}[h]
\centering
  \begin{subfigure}[t]{0.49\textwidth}
  		\centering
         \includegraphics[height=1.8in]{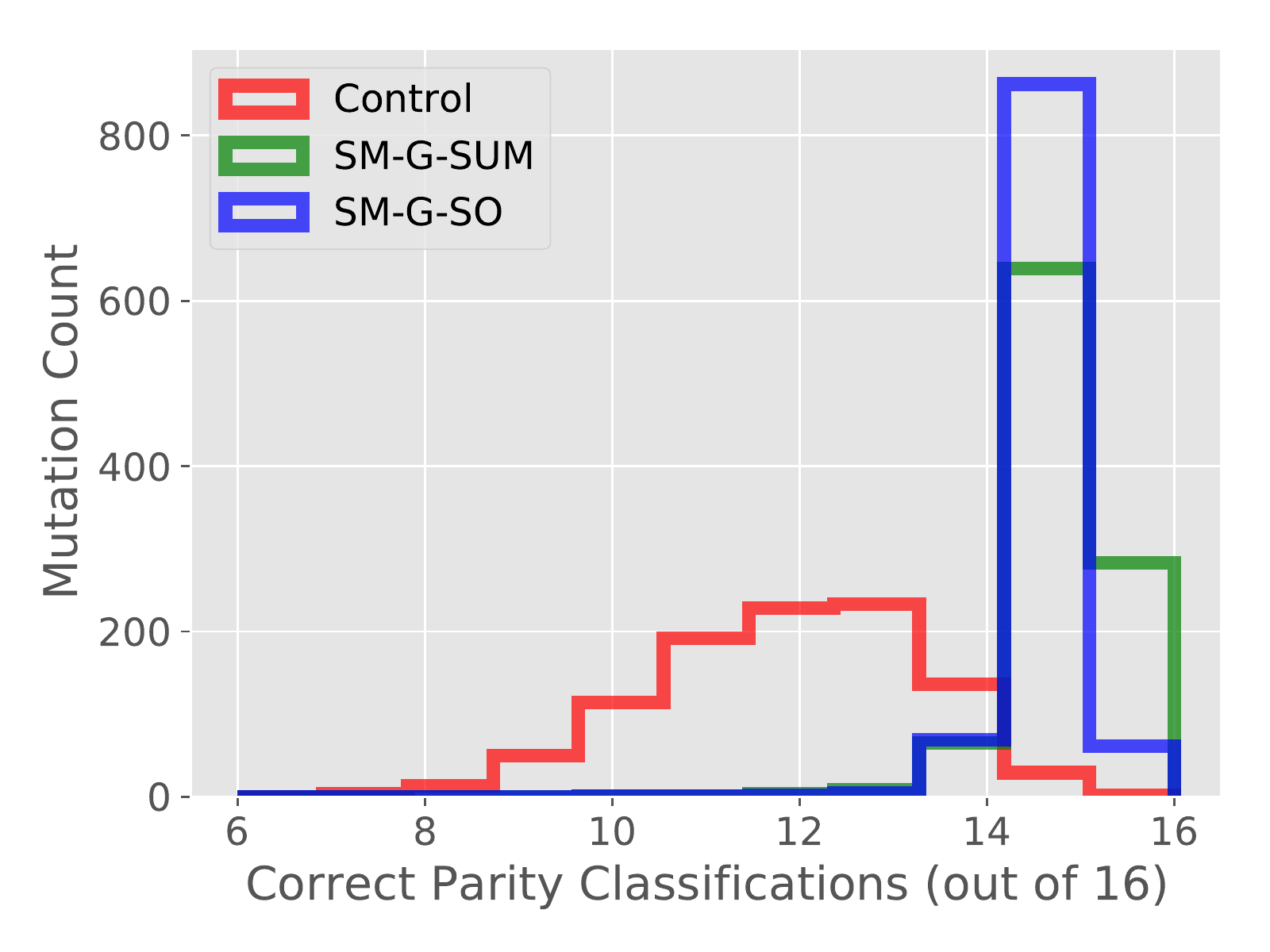}
        \caption{Representative Histogram}
    \end{subfigure}%
      \begin{subfigure}[t]{0.49\textwidth}
  		\centering
         \includegraphics[height=1.8in]{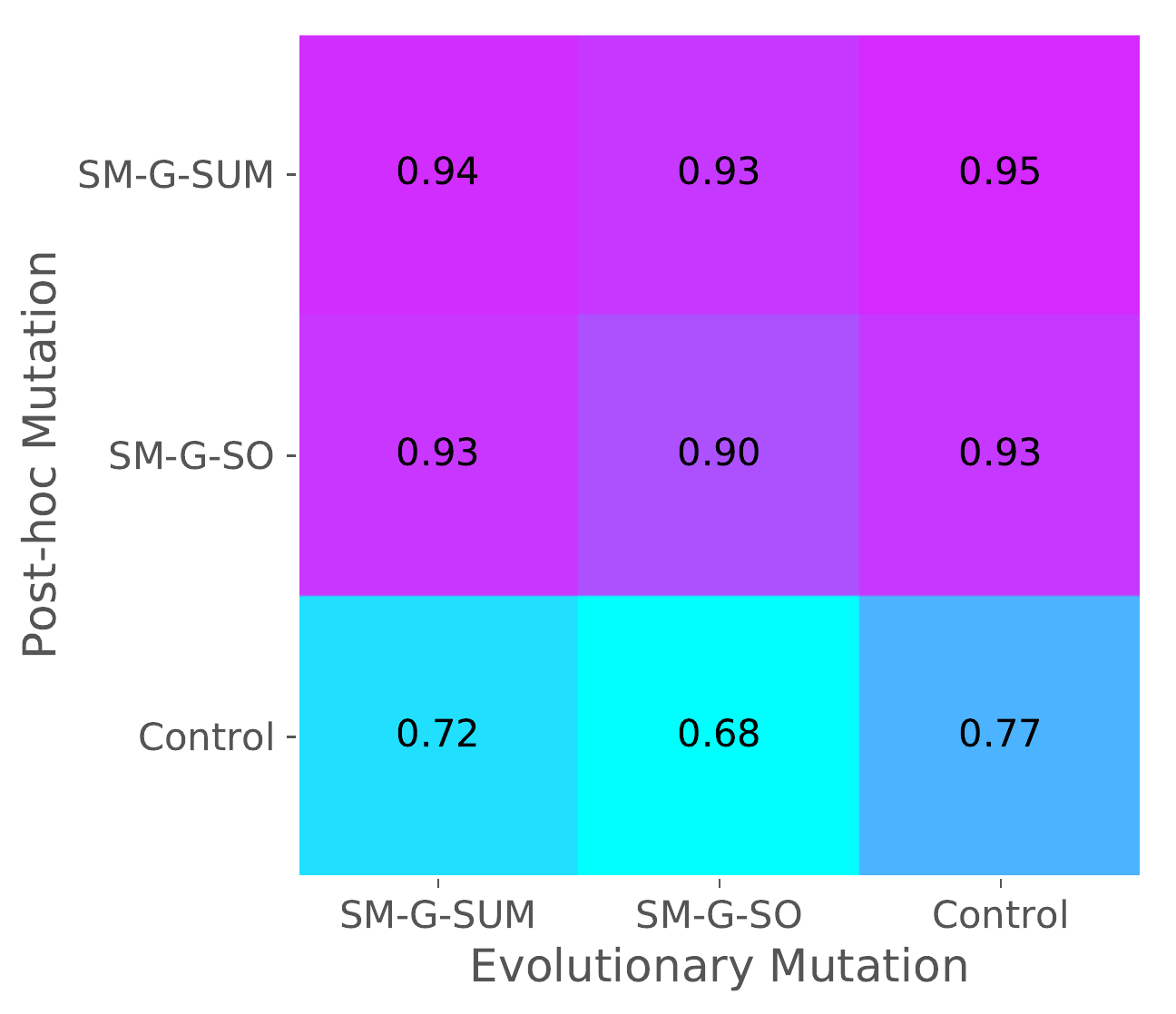}
        \caption{Comprehensive Comparison}
    \end{subfigure}%
  %\fbox{\rule[-.5cm]{0cm}{4cm} \rule[-.5cm]{4cm}{0cm}}
	\vspace{-0.1in}
  \caption{\label{fig:recurrent_safe}\textbf{Analyzing robustness of solutions in the Recurrent Parity task.} (a) For each mutation method, 1,000 perturbations by that method are generated from a representative solution (evolved by that method). The histogram of correct parity classifications (out of sixteen) achieved by perturbations are shown for SM-G-SUM, SM-G-SO, and the control. (b) The plot shows robustness to mutations averaged across solutions. For each evolutionary mutation method (i.e.\ the method generating the solution), the post-hoc Control mutation is significantly less robust than the post-hoc SM-G mutations (Mann-Whitney U-test; $p<0.001$).
 The conclusion is that SM-G methods do indeed produce safer mutations in this domain.}
\end{figure*}

%TODO: refactor this diagram
\begin{figure*}[h]
\begin{subfigure}[t]{0.32\textwidth}
  \centering
  \includegraphics[height=1.5in]{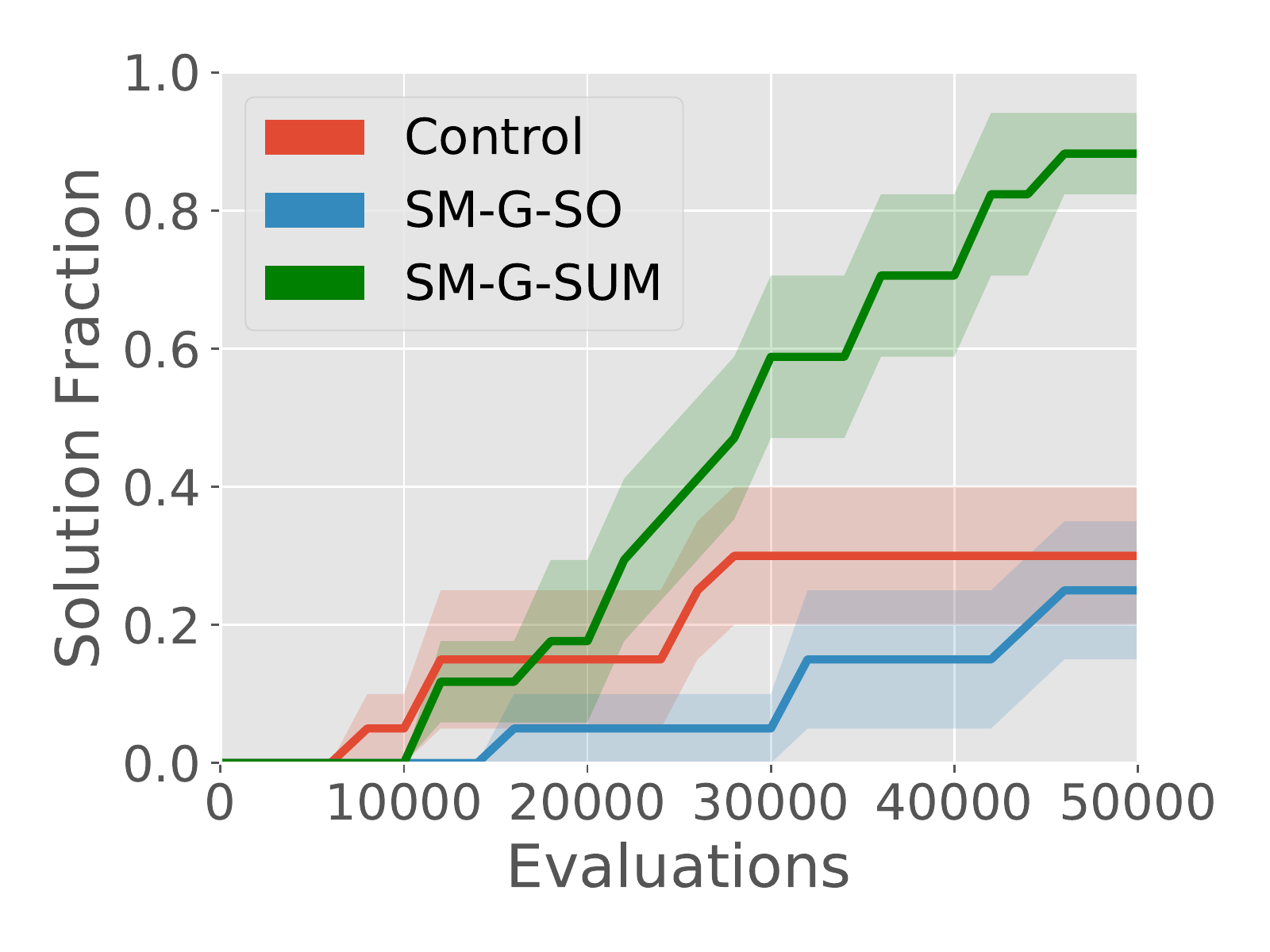}
  %\fbox{\rule[-.5cm]{0cm}{4cm} \rule[-.5cm]{4cm}{0cm}}
	\vspace{-0.125in}
  \caption{\label{fig:bigmaze}32 layers}	
  \end{subfigure}
\begin{subfigure}[t]{0.32\textwidth}
  \centering
  \includegraphics[height=1.5in]{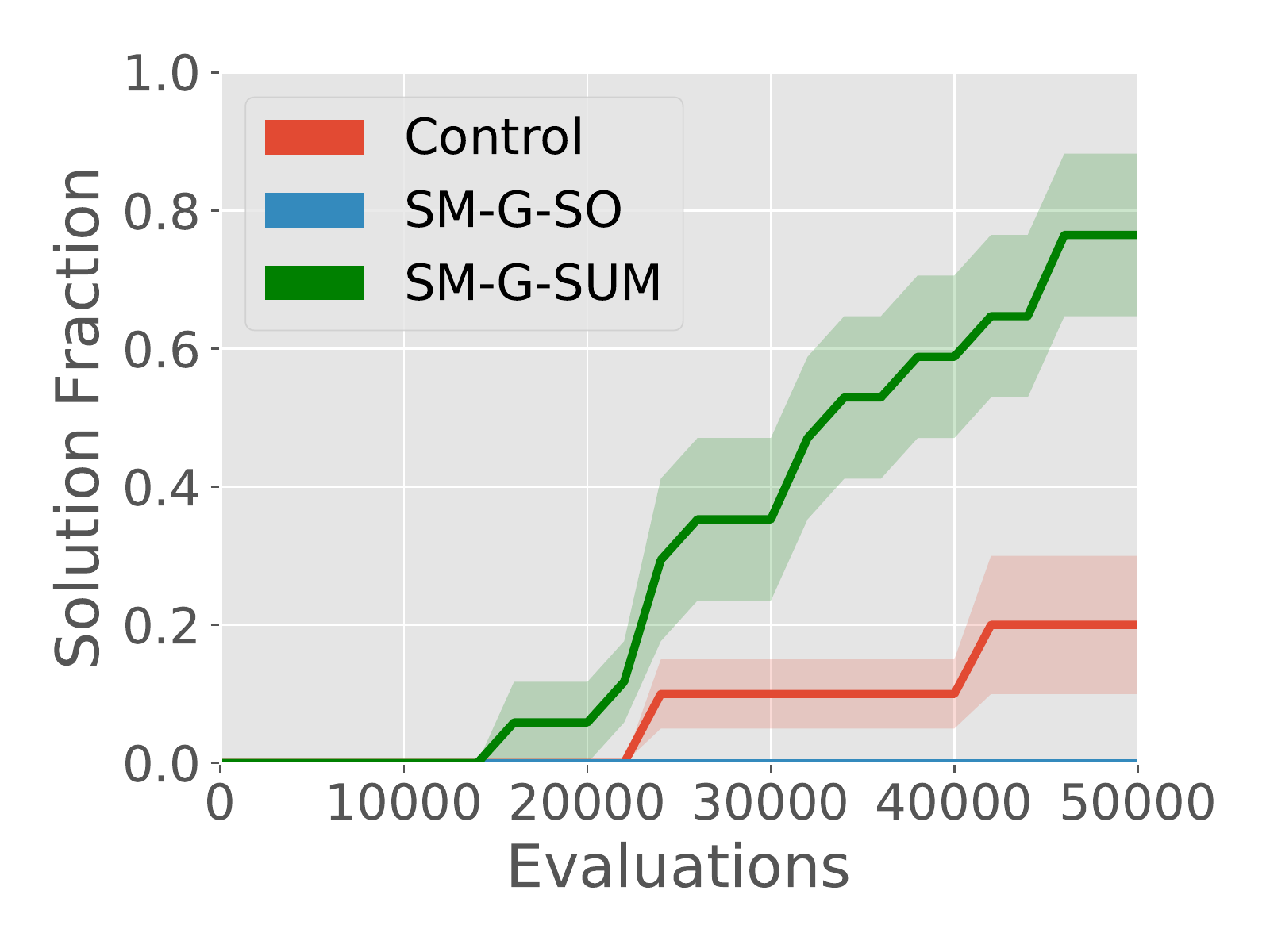}
  %\fbox{\rule[-.5cm]{0cm}{4cm} \rule[-.5cm]{4cm}{0cm}}
	\vspace{-0.125in}
  \caption{\label{fig:bigmaze}64 layers}	
\end{subfigure}
\begin{subfigure}[t]{0.32\textwidth}
  \centering
  \includegraphics[height=1.5in]{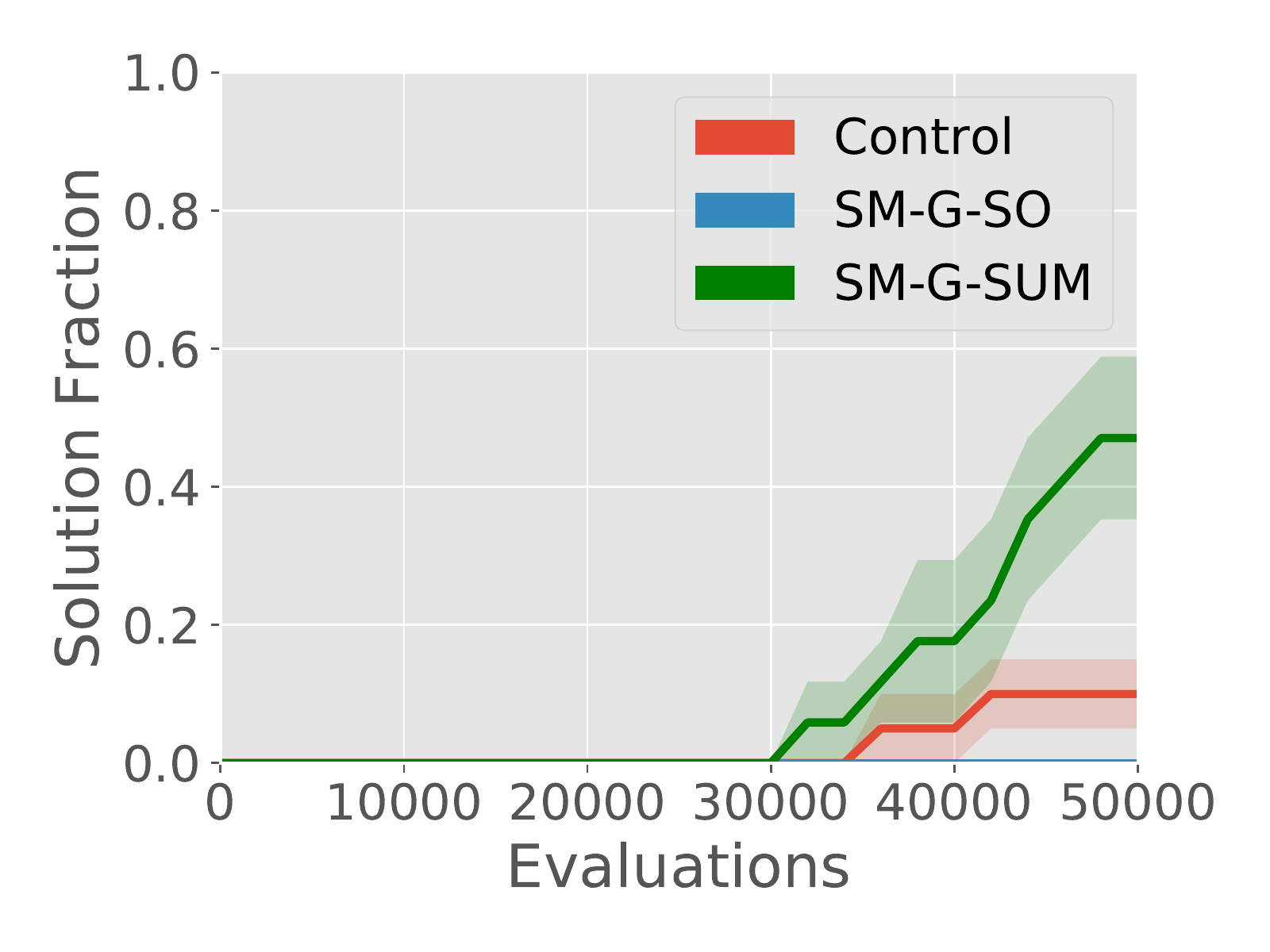}
  %\fbox{\rule[-.5cm]{0cm}{4cm} \rule[-.5cm]{4cm}{0cm}}
	\vspace{-0.125in}
  \caption{\label{fig:bigmaze}101 layers}	
\end{subfigure}
	\vspace{-0.1in}
\caption{\label{fig:bigmaze}\textbf{Performance comparison across evaluations on the large-scale NN task.} This expanded version of main-text figure 4 shows the fraction of solutions evolved by each method over increasing evaluations for 20 independent runs. SM-G-SUM evolves significantly more solutions than the standard mutation control and SM-G-SO in each of the (a) 32-layer, (b) 64-layer, and (c) 101-layer models.}	
\end{figure*}

\section{Hyperparameters}
For each experimental domain (besides those using the simple poorly-conditioned model), hyperparameter tuning was performed independently for each method. In particular, each method has a single hyperparameter corresponding to 
mutational intensity. For the control and the SM-G methods, this intensity factor is the standard deviation of the Gaussian
noise applied to the parent's parameter vector (which the SM-G methods then reshape based on sensitivity, but which the control leaves unchanged). In contrast, for SM-R, mutational intensity is given by the desired amount of divergence. 

Hyperparameter search was instantiated as a simple grid search spanning several orders of magnitude. In particular, each mutational method was evaluated for $8$ independent runs with each hyperparameter setting from the following set:
$\{1e-1,5e-2,1e-2,5e-3,1e-3,1e-4\}$. The best performing hyperparameter was then chosen based on highest average performance from the initial runs, and a final larger set of independent runs was conducted to generate the final results for each domain and method combination. These final hyperparameter settings are shown in table \ref{table:mutation}. Other hyperparameters (such as population size and tournament size) were fixed between methods and largely fixed between domains, and were subject to little exploration. These hyperparameters are shown in table \ref{table:evolution}.

%Toy domain
% Max evals: 2000
% Control 0.01
% SM-G-SUM 0.5
% SM-G-ABS 0.5
% SM-G-SO 0.5
% SM-R 0.5

\begin{table*}
\centering
\begin{tabular}{ l c c c c c}
  Domain & Control & SM-G-SUM & SM-G-ABS & SM-G-SO & SM-R \\ \hline
  Simple Model & 0.01 & 0.5 & 0.5 & 0.5 & 0.5 \\
  Recurrent Parity & 0.05 & 0.001 & 0.001 & 0.001 & 0.005 \\
  Breadcrumb Hard Maze & 0.05 & 0.1 & 0.005 & 0.01 & 0.005 \\
  Large-scale NN & 0.01 & 0.1 & & 0.01 &  \\
  Topdown Maze & 0.1 & 0.01 & & 0.01 &  \\
  First-person Maze & 0.005 & 0.01 & & 0.01 & 
  
%  7 & 8 & 9 \\
\end{tabular}
\vspace{0.1in}
\caption {\textbf{Mutation intensity settings for each domain and method combination.}\label{table:mutation}}
\end{table*}

\begin{table*}
\centering
\begin{tabular}{ l c c c c c}
  Domain & Population Size & Tournament Size & Maximum Evaluations \\ \hline
  Simple Model & 1 (hill-climber) & N\/A & 2k  \\
  Recurrent Parity & 250 & 5 & 100k  \\
  Breadcrumb Hard Maze & 250 & 5 & 100k \\
  Large-scale NN & 100 &5 & 50k  \\
  Topdown Maze & 250 & 5 & 50k &  \\
  First-person Maze & 250 & 5 & 50k &   
%  7 & 8 & 9 \\
\end{tabular}
\vspace{0.1in}
\caption {\textbf{Evolutionary hyperparameters across domains.}\label{table:evolution}}
\end{table*}

% Hyper-parameter search:
% Settings = [1e-1,5e-2,1e-2,5e-3,1e-3,1e-4]

%Recurrent Parity task
% Max evals: 100k
% Tournament size: 5
% Population size: 250

% Control 0.05
% SM-G-SUM 0.001
% SM-G-ABS 0.001
% SM-G-SO 0.001
% SM-R 0.005      

%Breadcrumb hard maze
% Max evals: 100k
% Tournament size: 5
% Population size: 250

% Control 0.05
% SM-G-SUM 0.1
% SM-G-ABS 0.005
% SM-G-SO 0.01
% SM-R 0.005      

%Big maze
% Max evals:  50k
% Population size: 100

% Control 0.01
% SM-G-SUM 0.1
% SM-G-SO 0.01

%TOPDOWN
% Max evals: 50k
% Population size: 250
% Control 0.1
% SM-G-SUM 0.01
% SM-G-SO 0.01
  
% RC maze
% Max evals: 50k 
% Population size: 250
% Control: 0.005
% SM-G-SUM 0.01
% SM-G-SO 0.01

%relevant hps
% pop size
% mutation power for each method
% tournament size
% max evals

\begin{comment}
\begin{center}
    \begin{tabular}{ | l | l | l | p{5cm} |}
    \hline
    Population Size &  & Max Temp & Summary \\ \hline
    Monday & 11C & 22C & A clear day with lots of sunshine.  
    However, the strong breeze will bring down the temperatures. \\ \hline
    Tuesday & 9C & 19C & Cloudy with rain, across many northern regions. Clear spells 
    across most of Scotland and Northern Ireland, 
    but rain reaching the far northwest. \\ \hline
    Wednesday & 10C & 21C & Rain will still linger for the morning. 
    Conditions will improve by early afternoon and continue 
    throughout the evening. \\
    \hline
    \end{tabular}
\end{center}
\end{comment}

\end{document}